\pdfoutput=1

\documentclass[11pt]{article}

\usepackage{acl}

\usepackage{times}
\usepackage{latexsym}

\usepackage[T1]{fontenc}

\usepackage[utf8]{inputenc}

\usepackage{microtype}
\usepackage{arydshln}
\usepackage{graphicx}
\usepackage{booktabs}
\usepackage{multirow}
\usepackage{microtype}
\usepackage{pifont}
\usepackage{array}
\usepackage{inconsolata}
\usepackage{adjustbox}
\usepackage{soul}
\usepackage{amsmath}
\usepackage{hyperref}
\usepackage{changepage}
\usepackage[inline]{enumitem}
\usepackage{makecell}

\newcolumntype{C}[1]{>{\centering\arraybackslash}p{#1}}
\newcommand{\RNum}[1]{\uppercase\expandafter{\romannumeral #1\relax}}

\newcommand\expertemoji{\raisebox{-2pt}{\includegraphics[width=0.9em]{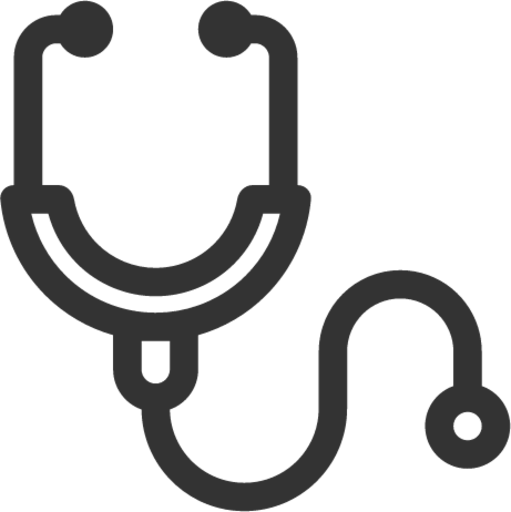}}}
\newcommand\expertemojitwo{\raisebox{-2pt}{\includegraphics[width=0.9em]{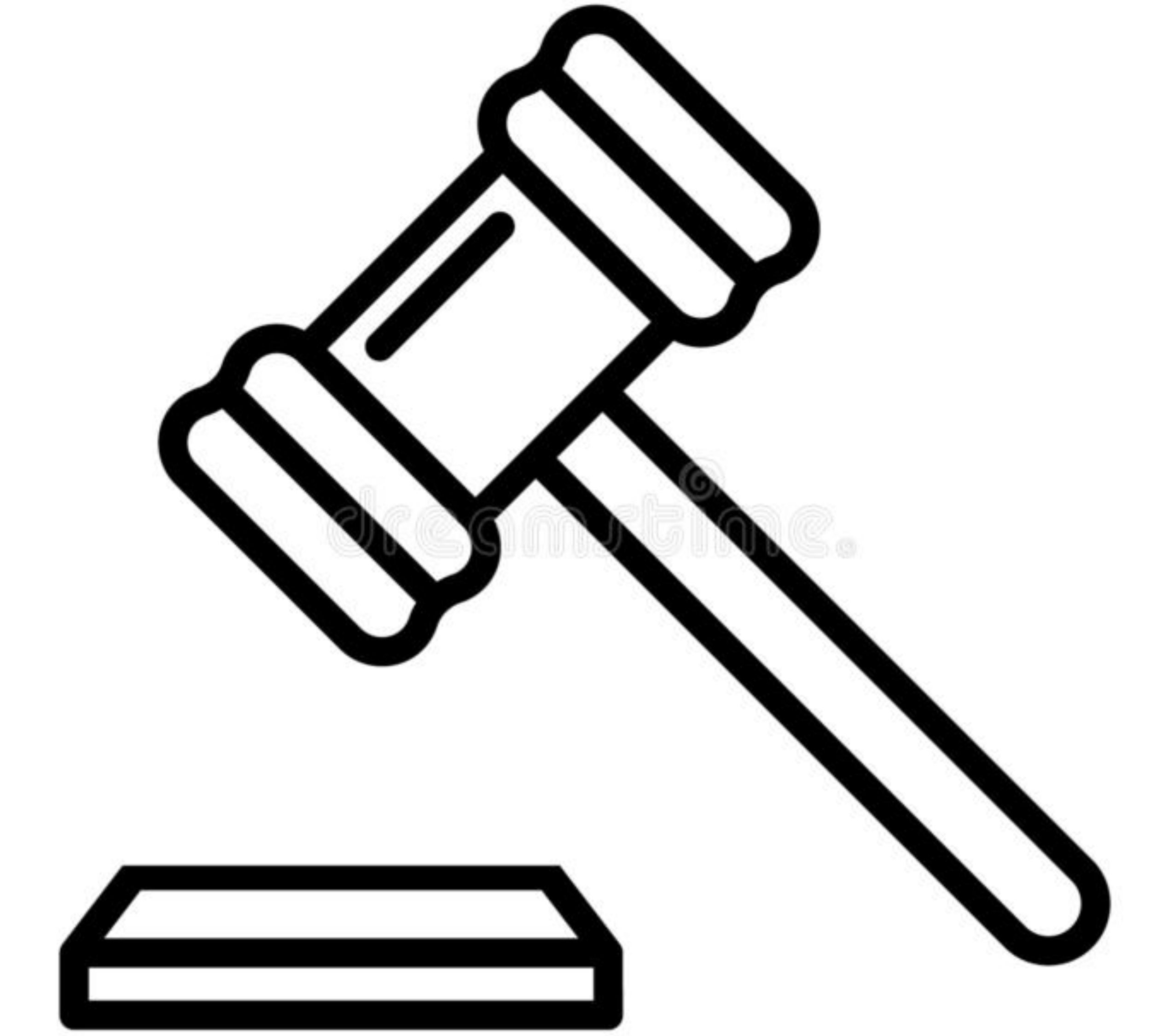}}}
\newcommand\robotemoji{\raisebox{-2pt}{\includegraphics[width=0.9em]{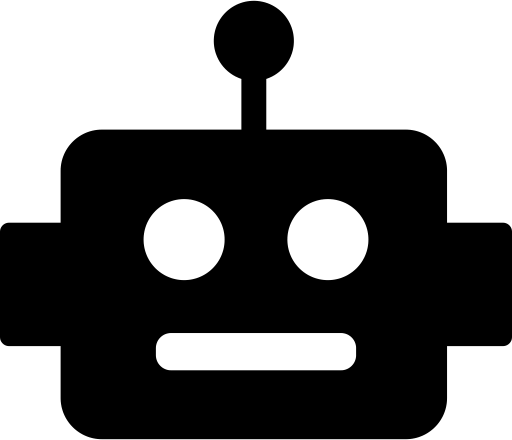}}}

\newcommand\expertqa{{\textsc{ExpertQA}}}

\title{\expertemoji{} \textsc{ExpertQA} \expertemojitwo{}: Expert-Curated Questions and Attributed Answers}

\author{
    Chaitanya Malaviya\textsuperscript{\robotemoji},
    Subin Lee\textsuperscript{\robotemoji},
    Sihao Chen\textsuperscript{\robotemoji},
    Elizabeth Sieber\textsuperscript{\expertemoji{}},\\
    \textbf{Mark Yatskar}\textsuperscript{\robotemoji},
    \textbf{Dan Roth}\textsuperscript{\robotemoji}
    \vspace{0.1in} \\
    \textsuperscript{\robotemoji}University of Pennsylvania 
    \textsuperscript{\expertemoji{}}University of Washington\\
    {\tt \{cmalaviy,subinlee,sihaoc,myatskar,danroth\}@upenn.edu}\\
    {\tt esieber@uw.edu}
}
\usepackage{pgfplots}
\pgfplotsset{compat=1.17}

\begin{document}
\maketitle

\begin{abstract}
As language models are adopted by a more sophisticated and diverse set of users, the importance of guaranteeing that they provide factually correct information supported by verifiable sources is critical across fields of study.
This is especially the case for high-stakes fields, such as medicine and law, where the risk of propagating false information is high and can lead to undesirable societal consequences.
Previous work studying attribution and factuality has not focused on analyzing these characteristics of language model outputs in domain-specific scenarios. 
In this work, we conduct human evaluation of responses from a few representative systems along various axes of attribution and factuality, by bringing domain experts in the loop.
Specifically, we collect expert-curated questions from 484 participants across 32 fields of study, and then ask the same experts to evaluate generated responses to their own questions.
In addition, we ask experts to improve upon responses from language models. The output of our analysis is \textsc{ExpertQA}, a high-quality long-form QA dataset with 2177 questions spanning 32 fields, along with verified answers and attributions for claims in the answers.\footnote{Code and dataset is available at \url{https://github.com/chaitanyamalaviya/ExpertQA}.}
\end{abstract}

\section{Introduction}

\begin{figure}[t!]
    \centering
    \includegraphics[width=\columnwidth,height=7cm,keepaspectratio]{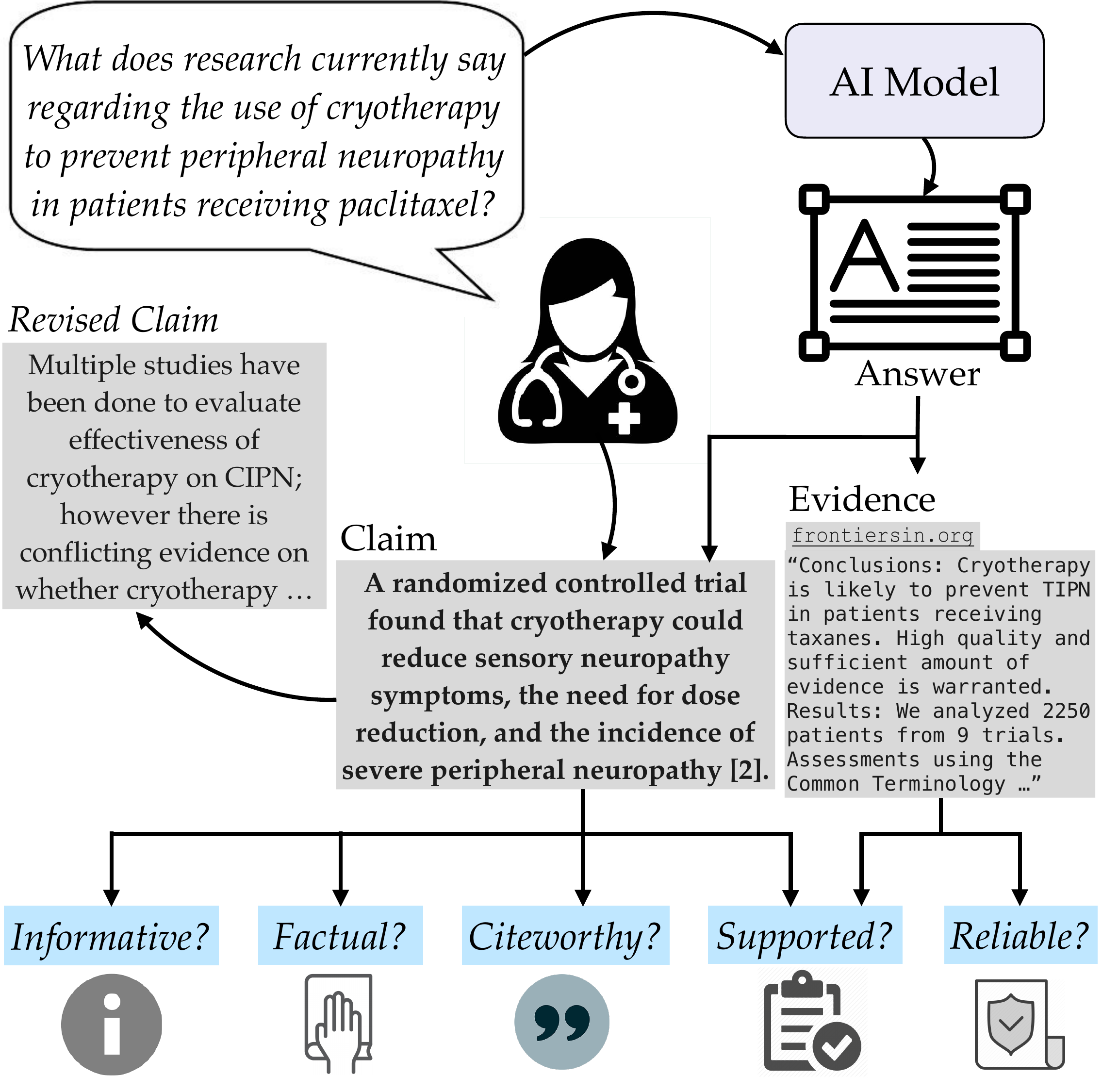}
    \caption{\textsc{ExpertQA} contains 2177 information-seeking questions formulated by experts spanning 32 fields, as well as expert-verified, model-generated answers to these questions. Each claim-evidence pair in an answer is judged by experts for various properties such as the claim's informativeness, factuality, citeworthiness, whether the claim is supported by the evidence, and reliability of the evidence source. Further, experts revise the original claims to ensure they are factual and supported by trustworthy sources.
    }
    \vspace{-10pt}
    \label{fig:expertqa}
\end{figure}

As the influence of large language models (LLMs) grows beyond the computer science community, 
\emph{experts} from various fields are rapidly adapting LLMs for assistance in information-seeking scenarios.
For example, medical professionals are using these systems for performing differential diagnosis \cite{lee2023benefits} and researchers are using them for faster literature surveys \cite{krenn2022,birhane2023,owens2023nature}.
While the use of LLMs in specialized domains has many potential benefits, it also carries significant risks.
False or hallucinated claims that are confidently phrased can potentially mislead experts and propagate societal harms, especially in high stakes domains such as medicine or law~\cite{evans2021truthful,dash2023evaluation,volokh2023libel,augenstein2023factuality}. 




Providing citations or attributions within generated responses is a promising direction for alleviating such concerns.
However, the quality of these attributions in model-generated responses, as well as the factuality of responses, is understudied in domain-specific settings. This is partly because we do not completely understand the specific information-seeking needs of experts.
Although experts from different fields are naturally best suited to aid with such an evaluation, expert evaluations are rarely conducted, as bringing experts in the loop can be time-consuming and costly.

To bridge this gap, we conduct an \textit{expert-in-the-loop} evaluation of attributed responses from a few representative systems. Having experts in the loop allows us to model a more realistic information-seeking scenario that helps us understand how people in different fields use LLMs and where their capabilities fall short. The output of our analysis is \textsc{ExpertQA}, a benchmark of information-seeking questions curated by experts from 32 fields, along with verified answers from representative systems. \textsc{ExpertQA} includes field-relevant questions, as well as claim-level judgements from experts along various axes of factuality and attribution. 



Our evaluation is conducted by first asking qualified experts to formulate questions from their field that they are curious about or have encountered in their professional lives~(\S\ref{sec:stage1}).
Responses to these questions are collected from a set of LLM-based systems that produce attributions for their answers (\S\ref{sec:systems}).
These include purely generative, retrieval-augmented, and post-hoc attribution systems.
We then ask experts to validate the claims and evidences found within responses to their own questions (\S\ref{sec:stage2}). 
Experts judge each claim for its informativeness to the question, its citeworthiness, and factuality.
They are also asked to judge how faithful the claim is to an accompanying evidence and rate the reliability of the evidence's source. 
Finally, experts revise each claim so it is faithful to reliable evidences and make a best effort attempt at ensuring the claim is factual. This overall process is described in Figure~\ref{fig:expertqa}.

Our findings (\S\ref{sec:analysis}) about representative systems from which responses are sampled suggest that:



\begin{enumerate}
    \item \textit{Retrieve-and-read systems generate more complete attributions compared to LLM prompting and post-hoc attribution, but struggle to produce citations for all cite-worthy claims.}
    \item \textit{The retrieval source significantly impacts the quality of attribution and overall factuality.} 
    \item \textit{High-stakes domains such as medicine and law suffer from a large percentage of incomplete attributions (35\% and 31\% incomplete attributions respectively) and many attributions come from unreliable sources (51\% attributions are not rated reliable by experts).}
\end{enumerate}



We also measure the extent to which existing automatic methods for attribution and factuality estimation \cite{bohnet2022attributed, min2023fact} correlate with expert judgements (\S\ref{sec:automatic}). We find that these metrics fall short in correlating with reference judgements of attribution and factuality. However, adapting these metrics to our data through finetuning results in improvements across domains.

The revised answers we collect can be used for improving and evaluating future models on long-form question answering. While similar datasets have been proposed~\cite{fan2019eli5}, examples in \textsc{ExpertQA} contain verified attributions and answers edited by experts.
We establish several baselines and show that we can improve models by finetuning on \textsc{ExpertQA} but that there is substantial room for improvement, both in terms of ROUGE and QAFactEval (\S\ref{sec:lfqa}).



\section{Expert-in-the-loop Evaluation}

\renewcommand{\arraystretch}{1.5} 

\begin{table*}[ht!]
\centering
\small
\scalebox{.8}{
\rowcolors{2}{gray!15}{white}
\begin{tabular}{ m{3.5cm} p{14cm} m{1cm} }
\rowcolor{gray!40}
\textbf{Field} & \textbf{Question} & \textbf{Types} \\ \toprule
 Anthropology & \textit{Why is it that Africa's representation is still a problem in modern day times regardless of the academic writings that state otherwise?} & \RNum{2},\RNum{7} \\
 Architecture & \textit{Suppose an architect decides to reuse an existing foundation of a demolished building, what is to be considered to ensure success of the project?} & \RNum{4} \\
 Biology & \textit{Can you explain the mechanisms by which habitat fragmentation affects biodiversity and ecosystem functioning, and provide examples of effective strategies for mitigating these impacts?} & \RNum{3},\RNum{6} \\
 Chemistry & \textit{Why does gallic acid have an affinity with trivalent iron ions?} & \RNum{1} \\
 Engineering \& Technology & \textit{How different will licensing a small modular reactor be as compared to licensing traditional large nuclear power plants?} & \RNum{7} \\
 Healthcare/Medicine & \textit{If a 48 year old woman is found to have an esophageal carcinoma that invades the muscularis propria and has regional lymph node metastases but no distant metastasis, what is her stage of cancer and what are possible recommended treatments?} & \RNum{1},\RNum{3} \\
 Law & \textit{Can direct evidence in a case that has been obtained illegally be considered by the court in some cases if it directly points to the defendant's guilt?} & \RNum{1} \\
 Music & \textit{What exercises would you do in a singing class with a teenager with puberphonia?} & \RNum{4} \\
 Physics \& Astronomy & \textit{Standard Model does not contain enough CP violating phenomena in order to explain baryon asymmetry. Suppose the existence of such phenomena. Can you propose a way to experimentally observe them?} & \RNum{5} \\
 Political Science & \textit{Despite the fact that IPCC was formed in 1988, several studies have showed that argubaly more than 50\% of all carbon emissions in history have been released since 1988. What does this show about IPCC and developed countries' efforts?} & \RNum{7} \\
 Visual Arts & \textit{Tell me the step by step process of recycling a canvas.} & \RNum{3} \\
 \bottomrule
\end{tabular}
}
\caption{Examples from \textsc{ExpertQA}. See Table~\ref{tab:full_examples} for a larger list showing an example from all fields. A large percentage of examples come from high-stakes fields such as Medicine and Law.}
\label{tab:examples}
\end{table*}

\renewcommand{\arraystretch}{1} 

\begin{table}[!ht]
\centering
\footnotesize
\scalebox{0.8}{
\begin{tabular}{c|C{6.5cm}|c}
    \toprule 
    & \textbf{Question Type} & \textbf{Count} \\
    \midrule
    \RNum{1} & Directed question that has a single unambiguous answer & 444 \\
    \RNum{2} & Open-ended question that is potentially ambiguous &  528 \\
    \RNum{3} & Summarization of information on a topic & 371 \\
    \RNum{4} & Advice or suggestions on how to approach a problem & 251 \\
    \RNum{5} & Question that describes a hypothetical scenario and asks a question based on this scenario & 853 \\
    \RNum{6} & Request for a list of resources where one can find more information & 160 \\
    \RNum{7} & Request for opinion on a topic & 207 \\
    \bottomrule
\end{tabular}}
\caption{Question types categorized according to various information needs that are part of \textsc{ExpertQA}.}
\label{tab:question_types}
\end{table}

The evaluation is conducted in multiple stages described below.
In the first stage, we ask experts to write questions from their field (\S\ref{sec:stage1}).
In the next stage, we present responses sampled from various systems back to the same experts for analysis (\S\ref{sec:stage2}). Further details about annotator backgrounds, costs and interfaces, are in Appendix~\ref{app:annotation}.

\subsection{Stage 1: Expert-Curated Questions}
\label{sec:stage1}

Participants are recruited through Prolific and are qualified as experts if they have i) received formal education, as well as, ii) at least 3 years of work experience in their field.
%
They are asked to write questions from their field which they have encountered in their professional life or ones they are genuinely curious about.
We ask them to formulate challenging technical questions, for which it may not be possible to find a single webpage that answers them completely. We note that this question collection is aimed at closely \textit{simulating} an information-seeking scenario with experts, since having access to real query logs is not feasible.

Each expert is asked to write 5 questions and to specify the question type(s) for each question (as shown in Table~\ref{tab:question_types}). These question types are formulated by adopting prior work that classifies information needs \cite{rose2004understanding}.
Because of their practical nature, at least two questions are required to be scenario-based questions (Type \RNum{5}, Table~\ref{tab:question_types}).
We collect questions 2177 questions from 524 experts in 32 fields, which are manually filtered for coherence and field-relevance. Examples of these questions are presented in Table~\ref{tab:examples}.


\begin{figure}[ht!]
    \centering
    \includegraphics[width=\columnwidth]{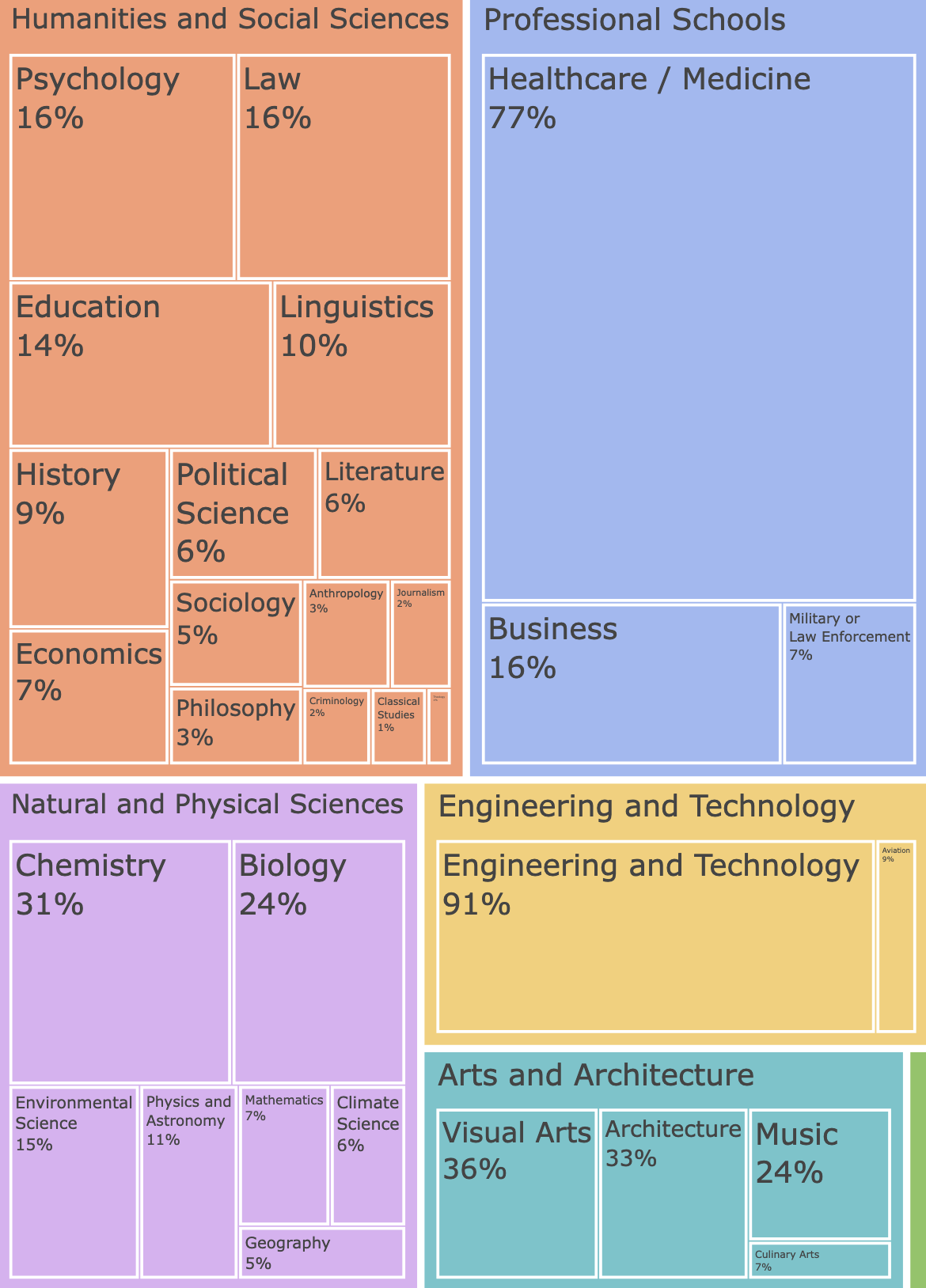}
    \caption{The distribution of questions across different fields in \textsc{ExpertQA}.}
    \label{fig:fields}
\end{figure}

\subsection{Stage 2: Answer and Claim Annotation}
\label{sec:stage2}

Next, we generate responses for the questions from stage 1 by prompting six different systems, described in \S\ref{sec:systems}, that provide attributions with their answers.
We split each answer into claims, where claims are considered at the granularity of a sentence and extracted using the spaCy sentence tokenizer \cite{spacy2}.\footnote{We also considered further increasing the atomicity of claims (like \citet{kamoi2023wice}) but evaluating finer-grained atomic claims incurs considerably higher annotation cost.}

In this stage of annotation, experts validate responses to their own questions on several dimensions of quality. 
92\% of annotators from stage 1 validated at least 1 of their own questions. 
The properties of answers and claims evaluated are shown in Table~\ref{tab:properties}.
Properties that judge answer quality are marked with \includegraphics[scale=0.04]{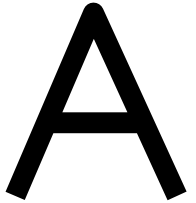} and those that judge evidence quality are marked with \includegraphics[scale=0.02]{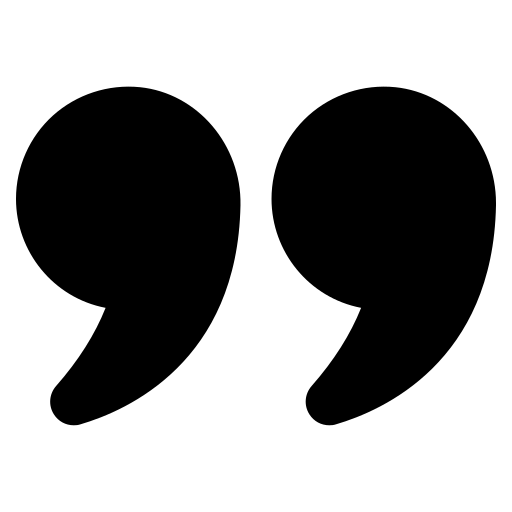}. After labeling these claim properties, annotators edit the response to ensure that the claim is factually correct and the given references support the claim.

\renewcommand{\arraystretch}{1.5} 

\begin{table*}[ht!]
\centering
\scalebox{.8}{
\rowcolors{2}{gray!15}{white}
\small
\begin{tabular}{ m{4cm} m{7.5cm} m{7cm} }
\rowcolor{gray!40}
\textbf{Property} & \textbf{Description} & \textbf{Ratings} \\ \toprule
 (\includegraphics[scale=0.04]{images/answer.png}) Answer Usefulness & \textit{Is the answer useful in responding to the question?} & \{\textit{Useful}, \textit{Partially useful}, \textit{Not useful at all}\} \\
 (\includegraphics[scale=0.04]{images/answer.png} + \includegraphics[scale=0.02]{images/quote.png}) Attribution & \textit{Is the claim supported by its accompanying evidence?} & \{\textit{Complete}, \textit{Partial} or \textit{Incomplete}, \textit{Missing}, \textit{N/A} (if link broken)\} \\
 (\includegraphics[scale=0.04]{images/answer.png}) Informativeness & \textit{Is the claim relevant to answering the question?} & \{\textit{Very relevant}, \textit{A bit relevant}, \textit{Not too important}, \textit{Uninformative}\} \\
 (\includegraphics[scale=0.04]{images/answer.png}) Factuality & \textit{Is every word of the claim factually correct?} & \{\textit{Definitely correct}, \textit{Probably correct}, \textit{Unsure}, \textit{Likely incorrect}, \textit{Definitely incorrect}\} \\
 (\includegraphics[scale=0.02]{images/quote.png}) Source Reliability & \textit{Is the accompanying evidence (if any) for the claim found on a website you would consider reliable?} & \{\textit{Reliable}, \textit{Somewhat Reliable}, \textit{Not reliable at all}\} \\
 (\includegraphics[scale=0.04]{images/answer.png}) Cite-worthiness &\textit{Is the claim necessary to be cited?} & \{\textit{Yes}, \textit{No}\} \\
 \bottomrule
\end{tabular}
}
\caption{Properties of claims and evidences annotated in \textsc{ExpertQA}.}
\vspace{-5pt}
\label{tab:properties}
\end{table*}
\renewcommand{\arraystretch}{1} 

\section{Systems Evaluated}
\label{sec:systems}
We now describe the classes of systems from which we sampled responses to questions. 
All systems we evaluated produce an answer string and attributions in the form of in-line citations. Attributions are returned as URLs or passages along with URLs from where they are retrieved. Experimental details such as prompts are in Appendix~\ref{app:experimental}.

\paragraph{LLM as generator + retriever.} In this paradigm, we prompt large language models in a closed-book fashion \cite{brown2020language, OpenAI2023GPT4TR} to generate an answer with in-line citations where they provide URLs for each citation. This means that the model essentially has to generate a URL from its parametric memory. We consider GPT-4 as the LLM from which we sample responses (\texttt{gpt4}).

\paragraph{Post-hoc retrieval.} This system differs from the above, as we only prompt LLMs to generate answers without attribution, and perform retrieval of evidence for a claim as a post-hoc step. This renders the attributions naturally unfaithful, but we believe this is still a worthwhile approach to investigate because of the strength of LLMs as generators and retrievers independently. The attribution corpora we consider are Sphere \cite{sphere} (\texttt{post\_hoc\_sphere\_gpt4}), which is a large static dump of CommonCrawl, and Google search results (\texttt{post\_hoc\_gs\_gpt4}).

\paragraph{Retrieve-and-read.} In this class of systems, we first retrieve evidence for a question and then prompt a model to use the retrieved evidence to answer the question \cite{chen-etal-2017-reading}. As our attribution corpus, we again consider Sphere \cite{sphere} (\texttt{rr\_sphere\_gpt4}) and Google search results (\texttt{rr\_gs\_gpt4}). We use BM25 \cite{robertson2009probabilistic} for retrieving from Sphere. We then generate an answer using GPT-4, providing the retrieved evidence as context. The model is instructed to generate in-line citations for each sentence, which refer to the passages in the context.

\paragraph{Commercial.} We also consider commercial systems such as BingChat.\footnote{The precise implementation of these systems is proprietary, but we can still draw conclusions about their utility.}
We sample responses using the balanced mode of BingChat (\texttt{bing\_chat}).

\begin{table}[!t]
\centering
\scalebox{0.85}{
\begin{tabular}{l|c|c}
    \textbf{System} & \textbf{Count} & \textbf{Abstention Rate} \\ \toprule
    \texttt{gpt4} & 174 & 0\% \\
    \texttt{bing\_chat} & 470 & 0.01\% \\
    \texttt{rr\_sphere\_gpt4} & 279 & 37.89\% \\
    \texttt{rr\_gs\_gpt4} & 452 & 22.69\% \\
    \texttt{post\_hoc\_sphere\_gpt4} & 403 & 0\% \\
    \texttt{post\_hoc\_gs\_gpt4} & 399 & 0\% \\
    \bottomrule
\end{tabular}}
\caption{Number of examples sampled from different systems and the abstention rates of different systems.}
\vspace{-10pt}
\label{tab:system_freq}
\end{table}

\subsection{Response Sampling}

We sample uniformly from all systems but exclude abstained answers and constrain each answer to contain at most 10 claims.
Attributions from \texttt{gpt4} often point to broken links, so we sampled more responses from the other systems.
The number of examples from each system and how frequently they abstain
are reported in Table~\ref{tab:system_freq}.

\section{Analysis}
\label{sec:analysis}

\subsection{Data Statistics}

\begin{figure}[ht]
    \centering
    \includegraphics[width=\columnwidth,height=9cm,keepaspectratio]{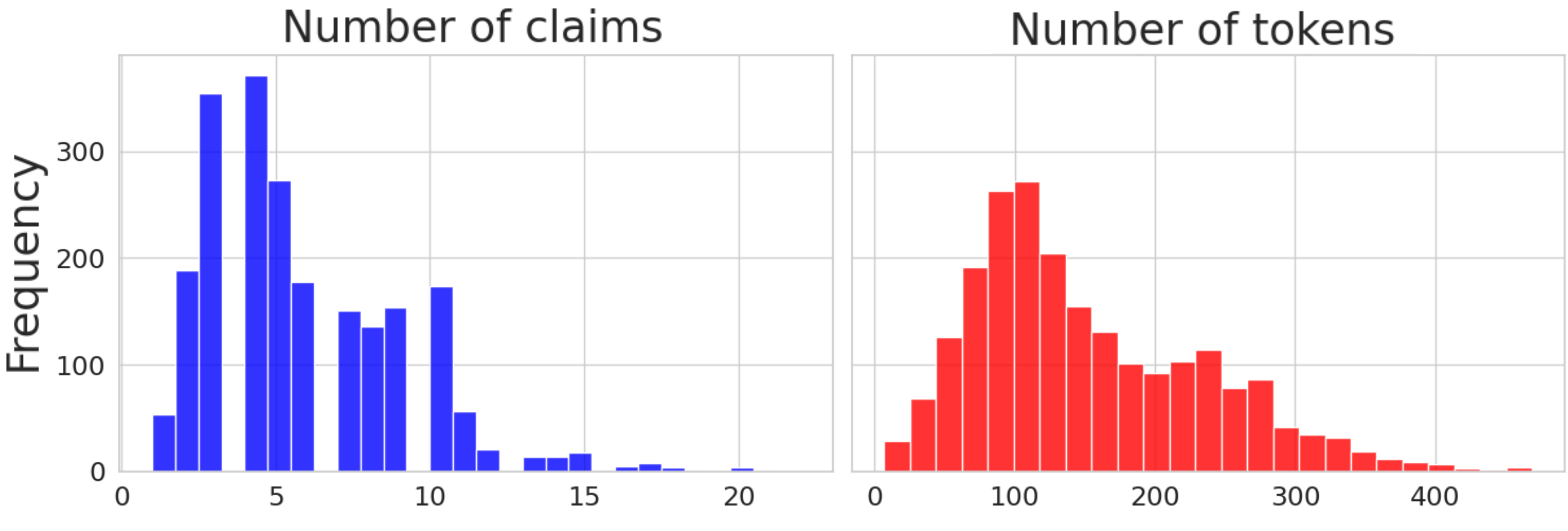}
    \caption{Histogram of the number of claims and number of tokens across all examples in \textsc{ExpertQA}. The average number of claims and tokens across examples is 5.79 and 152.12 respectively}
    \vspace{-10pt}
    \label{fig:lengths}
\end{figure}

The total number of examples validated in \textsc{ExpertQA} is 2177. The distribution of the number of claims and tokens is shown in Figure~\ref{fig:lengths}. The distribution of examples across fields and question types are presented in Figure~\ref{fig:fields} and Table~\ref{tab:question_types} respectively.


\subsection{Manual Analysis}
\label{sec:manual}

\begin{figure}[h]
    \centering
    \includegraphics[width=\columnwidth,height=9cm,keepaspectratio]{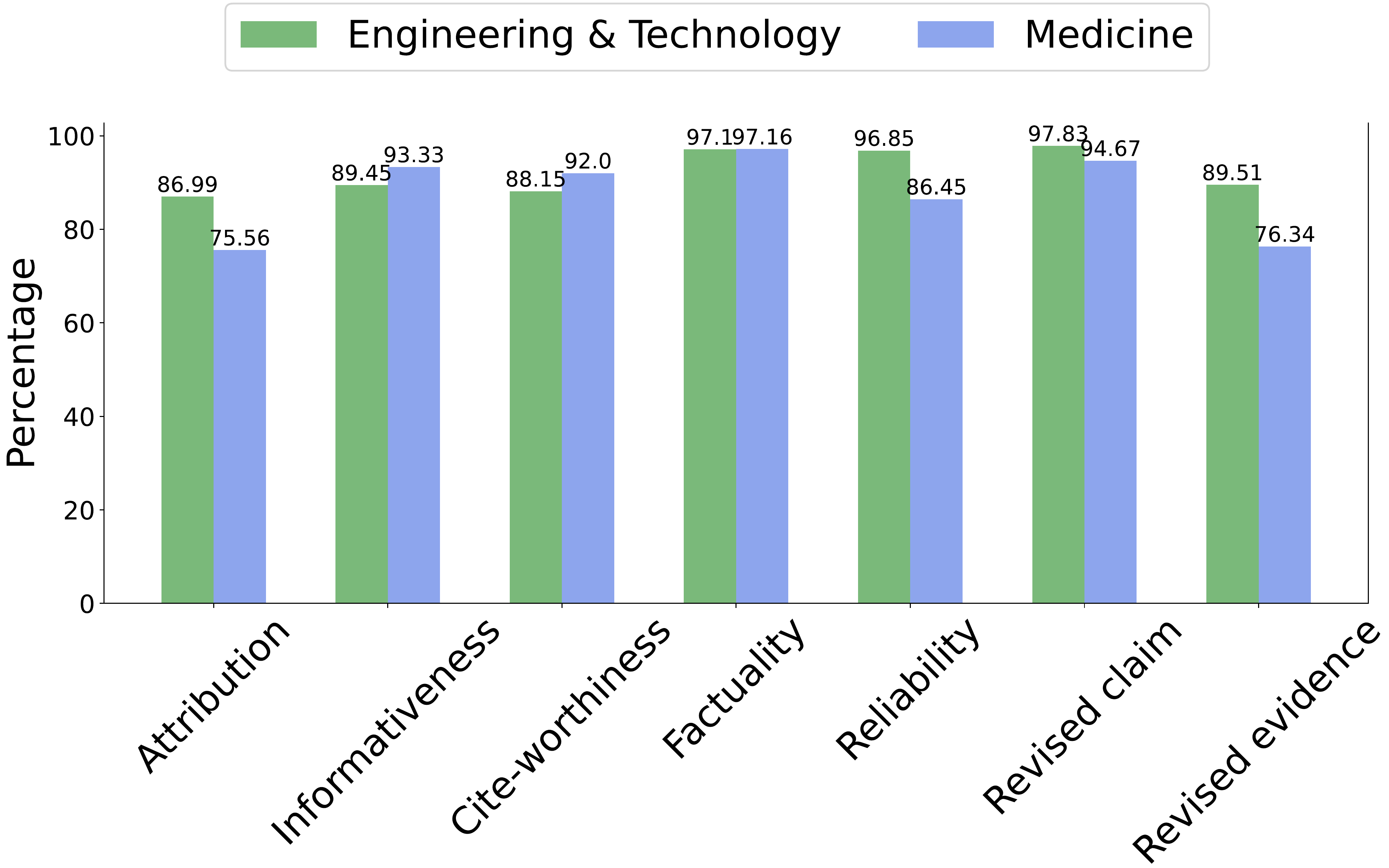}
    \caption{Percentage agreement on claim annotations based on our manual analysis.}
    \label{fig:manual_analysis}
\end{figure}

To estimate the reliability of the collected human labels, we, the authors, computed our agreement with the reference labels from two fields in which the authors are experts. We sampled 60 questions each from Engineering \& Technology and Medicine, sampling answers uniformly from all systems.
For each claim, we label our agreement with the reference label for each property from Table~\ref{tab:properties}.
Our analysis, as summarized in Figure~\ref{fig:manual_analysis}, shows high agreement (> 85\%) for most labels in both fields considered.

\begin{figure*}[t!]
    \centering
    \includegraphics[width=2\columnwidth]{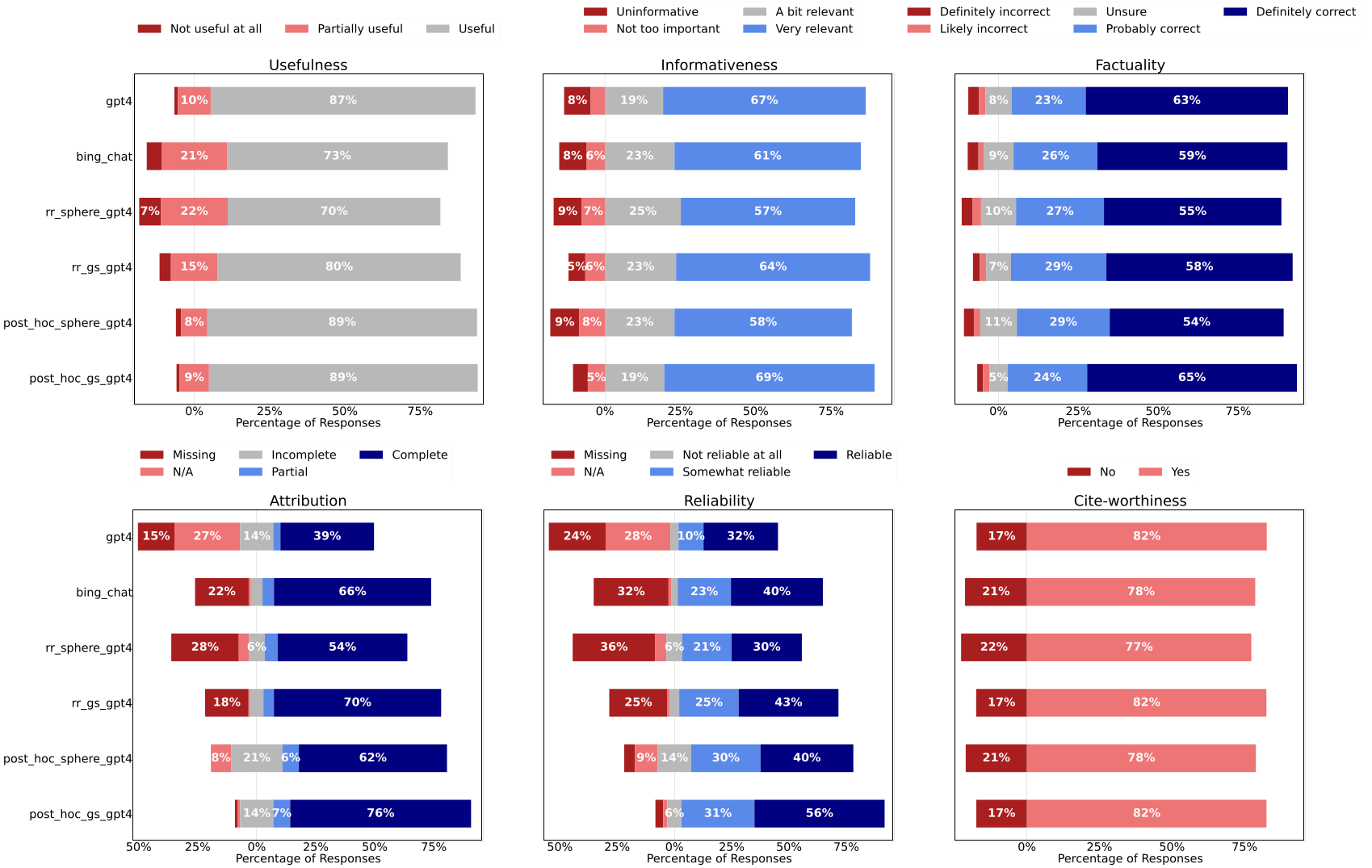}
    \caption{The Likert distribution of labels for the different properties of answers / claims, annotated by experts. The top 3 properties (answer usefulness, claim informativeness and factuality) are  judgements of answer quality and the bottom 3 (claim/evidence attribution, source reliability and claim cite-worthiness) are attribution quality.}
    \vspace{-10pt}
    \label{fig:likert}
\end{figure*}

\subsection{Analysis of Expert Evaluations}

We present the Likert distribution for claims across all systems and properties in Figure~\ref{fig:likert}.
Below we summarize the main conclusions from our analysis.

\paragraph{Majority of answers are useful, but answers from purely generative systems are considered more useful.} We find that $\sim$87-89\% of answers from \texttt{gpt4} are marked useful. The retrieve-and-read systems (as well as \texttt{bing\_chat}) are marked slightly less useful (73-80\%), likely because retrieved evidences are not always highly relevant. Choosing relevant evidences from the web using Google search results in more useful answers than with the smaller Sphere corpus. Analyzing responses marked not useful, we find that \textbf{systems struggle with targeted responses to long-tail queries, by resorting to patterns such as hedging, or providing generic or vague information}.

\paragraph{Retrieve-and-read systems often generate complete attributions, but struggle to produce citations for all cite-worthy claims.} While these systems have a stronger inductive bias to use the retrieved evidence to generate a response, they do not always produce attributions for cite-worthy claims (18\% of these claims are missing attributions)\footnote{Figure~\ref{fig:likert} shows the Likert distribution of attribution labels on those claims deemed cite-worthy by experts.}. On the other hand, post-hoc attribution systems return attributions for every single claim by design, but return more incomplete attributions. 
Lack of context during post-hoc retrieval can be an issue for retrieving valid attributions.

Finally, without retrieval, while \texttt{gpt4} generates citations to plausible domains (for e.g., \texttt{nasa.gov} for astronomy, \texttt{nih.gov} for medical claims), the content on these webpages is usually totally mismatched (more than 60\% of the time). Across systems, we find that \textbf{because domain-specific claims are long-tail and niche, it is hard to find reliable evidence on web documents that completely supports such claims.}

\paragraph{Both vanilla prompting and retrieval-augmented systems generate mostly \textit{very relevant} claims to the question.} At the same time, a significant percentage of claims (30-40) are not very relevant. This includes void claims (that simply restate the question or state simplistic facts). This suggests that there is a lot of room in making answers concise and relevant.

\paragraph{Just over half the claims are labeled as \textit{definitely correct} by experts.} While a significant percentage of claims are labeled as correct (\textit{probably} or \textit{definitely}), experts do not instill high confidence in the factual correctness of claims. This might be because it is hard to judge factuality with a high degree of confidence in a short time frame. 
Once again, a smaller retrieval corpus (\texttt{rr\_sphere\_gpt4}) results in less factual claims as the model may be more likely to hallucinate.

\paragraph{The retrieval corpus has a significant effect on expert judgements of source reliability.} Expert judgements of source reliability are influenced by the corpus from which evidences are retrieved. Corpora such as Sphere contain evidences that are unreliable to experts (for both \texttt{rr\_sphere\_gpt4} and \texttt{post\_hoc\_sphere\_gpt4}). Note also that we do not account for the authoritativeness of domains when retrieving from Sphere. For example, in a question about breast cancer, evidence from a comment on a blog is retrieved and is naturally judged unreliable. Using Google search improves reliability judgements significantly.

\paragraph{Majority of claims are deemed cite-worthy across systems.} Only around 17-22\% claims are judged not citeworthy by the experts. This suggests that most claims in responses to expert-curated questions warrant providing supporting evidence. 


\paragraph{Domain and Question Type Trends.} Figure~\ref{fig:stats_by_field} shows the distribution of labels across fields. 
The percentage of claims labeled factually correct is fairly high (>85\%) for many fields. However, we note that across all annotated claims, \textbf{high-stakes domains such as medicine and law suffer from a significant percentage of incomplete attributions} (around 35\% and 31\% unsupported claims respectively). Further, \textbf{a large percentage of claims present evidences from unreliable sources} (for eg, $\sim$51\% of medical claims have attributions from sources that are not \textit{Reliable}). The trends across question types (Figure~\ref{fig:stats_by_qtype}), systems clearly struggle with Type \RNum{6} questions that request for a list of resources, as claims are less informative, factual, and supported by evidence. 








\section{Automatic Estimation of Attribution and Factuality}
\label{sec:automatic}

Prior work has proposed automatic methods to predict attribution and factuality of claims. We evaluate how reliably these methods reflect the expert labels in our collected data.
We evaluate the effectiveness of these methods for claims in \expertqa.  In both cases, we observe that \textbf{current methods show high precision but low recall when compared with human judgements.}. 

\subsection{Automatic Attribution Estimation} 
Under the \textit{attributable to identifiable sources} (AIS) framework of \citet{rashkin2021measuring}, previous work has found NLI models to be effective in providing automated AIS (AutoAIS) estimates \cite{bohnet2022attributed}.
\begin{table}[!t]
\centering
\scalebox{0.8}{
\begin{tabular}{l|c|c}
    \textbf{System} & \textbf{AutoAIS} & \textbf{Num. Claims}  \\ \toprule 
    \texttt{gpt4} & .156 & 149  \\
    \texttt{bing\_chat} & .320 & 992  \\
    \texttt{rr\_sphere\_gpt4} & .689 & 732 \\
    \texttt{rr\_gs\_gpt4} & .778 & 1415 \\
    \texttt{post\_hoc\_sphere\_gpt4} & .281 & 1158  \\
    \texttt{post\_hoc\_gs\_gpt4} & .241 & 1500 \\    \bottomrule
\end{tabular}}
\caption{AutoAIS score (more attributable$\rightarrow$1, less attributable$\rightarrow$0) of predicted responses by the systems. Only claims annotated as \emph{citeworthy} and \emph{with complete support} are considered.}
\vspace{-10pt}
\label{tab:autoais-results}
\end{table}

Following previous work, we use an NLI model \cite{honovich-etal-2022-true-evaluating} to predict binary attribution labels of claim-evidence pairs in \textsc{ExpertQA}. 
For evidences longer than the model's sequence length (512), we use the stretching technique from \citet{schuster-etal-2022-stretching}, where we split the evidence into sentences and use the top-2 sentences with highest entailment scores as evidence. 

Table~\ref{tab:autoais-results} shows the macro-averaged AutoAIS scores for the claims annotated as having complete attributions. 
Compared to human judgments, the AutoAIS scores show large variance across systems. Notably, attributions from post-hoc retrieval systems receive much lower AutoAIS scores compared to retrieve-and-read systems.

\begin{table}[!t]
\centering
\scalebox{0.8}{
\begin{tabular}{l|ccc|ccc}
    \multirow{2}{*}{\textbf{System}} & \multicolumn{3}{c|}{zero-shot} & \multicolumn{3}{c}{finetuned} \\
     & \textbf{P} & \textbf{R} & \textbf{F1} & \textbf{P} & \textbf{R} & \textbf{F1} \\ \toprule 
    \texttt{gpt4} & .33 & .02 & .05 & .52 & .32 & .39 \\
    \texttt{bing\_chat} & .97 & .26 & .41 & .90 & .90 & .90 \\
    \texttt{rr\_sphere\_gpt4} & .89 & .59 & .71 & .83 & .90 & .87 \\
    \texttt{rr\_gs\_gpt4} & .86 & .74 & .79 & .87 & .98 & .92\\
    \texttt{post\_hoc\_sphere\_gpt4} & .92 & .28 & .43 & .79 & .97 & .87\\
    \texttt{post\_hoc\_gs\_gpt4} & .87 & .17 & .29 & .77 & .95 & .85 \\ \midrule
    \texttt{all} & .88 & .38 & .53 & .82 & .91 & .86 \\
    \bottomrule
\end{tabular}}
\caption{Precision, Recall and F1 scores of AutoAIS labels predicted by the TRUE NLI model (0-shot vs. finetuned version on the ExpertQA train split) against human attribution judgements in \textsc{ExpertQA}.}
\vspace{-3pt}
\label{tab:autoais-human}
\end{table}

\begin{table}[!ht]
\small
\centering
\begin{tabular}{p{\linewidth}}
    \toprule
    \rowcolor{gray!15}
    \textbf{Error Type}: \textit{Fine-grained Information Sensitivity}\\
    \midrule
    \textbf{Claim} (\texttt{post\_hoc\_sphere\_gpt4}): For water with a low pH (acidic), you can add a base or alkaline compound, such as baking soda (sodium bicarbonate) or \textcolor{red}{calcium carbonate}, to raise the pH [1].  \\
    \textbf{Attribution [1]}: 
    ... To raise or lower pH, a pool custodian simply adds acids or alkalis into the water. For example, adding sodium carbonate (soda ash) or sodium bicarbonate (baking soda) will generally raise the pH and adding muriatic acid or sodium bisulfate will lower the pH. \\  
    \textbf{Human}: \textit{Cite-Worthy \& Complete Support} \\ 
    \textbf{AutoAIS}: \textit{0 (No or Partial Support)}. \\
    \midrule 
    \rowcolor{gray!15}
    \textbf{Error Type}: \textit{Multi-Source Attributions}\\
    \midrule
    \textbf{Claim} (\texttt{bing\_chat}): Other radiological signs of fetal death include \textcolor{orange}{gas in the fetus} or in the portal and umbilical vessels [1], and \textcolor{blue}{Deuel's halo sign} [2]. \\
    \textbf{Attribution [1]}: ... \textcolor{orange}{Intrafetal gas} is an unequivocal sign of fetal death provided it can be conclusively differentiated from maternal gas, shadows. ... \\
    \textbf{Attribution [2]}: Radiological investigation is warranted in the antenatal patient only if the findings are likely to influence future management. The major radiological signs of fetal death include overlapping of the cranial bones and \textcolor{blue}{Deuel's halo sign} \\
    \textbf{Human}: \textit{Cite-Worthy \& Complete Support} \\ 
    \textbf{AutoAIS}: \textit{0 (No or Partial Support)}. \\
    \bottomrule
\end{tabular}
\caption{Examples of typical errors of AutoAIS against human judgements in \textsc{ExpertQA}.}
\vspace{-10pt}
\label{tab:ais-example}
\end{table}

We compare the per-claim AutoAIS predictions to human judgements of attribution in Table~\ref{tab:autoais-human}. 
The results suggest that AutoAIS estimates have high-precision yet low-recall against human judgements of attribution. 
To understand the discrepancy between NLI model behavior vs. human judgements, we highlight a few typical examples of attribution errors in Table~\ref{tab:ais-example}. 
For NLI models, every part of the claim needs to be verifiable with the evidence, but human judgements involve more implicit world knowledge, e.g. \emph{calcium carbonate is an alkali}. Another common mistake involves synthesizing information from multiple evidences. We observe multi-source attributions to be particularly common among \texttt{bing\_chat} and retrieve-and-read systems.  

\subsection{Automatic Factuality Estimation} 
Prior work has proposed methods \cite{manakul2023selfcheckgpt,min2023fact} to estimate the factuality of model generations.
In particular, we use FActScore \cite{min2023fact} to estimate factuality of claims.
We first break down each claim into fine-grained atomic claims using few-shot prompting with \texttt{text-davinci-003}. 
We then retrieve the top-3 relevant passages using Google search with the atomic claim as the query. The atomic claim and the evidence passages are then used to prompt \texttt{gpt-3.5-turbo} to say whether the atomic claim is \textit{True} or \textit{False}. The FActScore of a claim is the FActScore averaged across its atomic claims.

\begin{table}[t]
\centering
\scalebox{0.8}{
\begin{tabular}{l|c|c|c}
    \textbf{System} & \textbf{F1 (T)} & \textbf{F1 (F)} & \textbf{F1 (overall)} \\ \toprule 
    \texttt{gpt4} & 0.919 & 0.108 & 0.852 \\
    \texttt{bing\_chat} & 0.912 & 0.134 & 0.841 \\
    \texttt{rr\_sphere\_gpt4} & 0.884 & 0.106 & 0.795 \\
    \texttt{rr\_gs\_gpt4} & 0.927 & 0.068 & 0.865 \\
    \texttt{post\_hoc\_sphere\_gpt4} & 0.898 & 0.132 & 0.817 \\
    \texttt{post\_hoc\_gs\_gpt4} & 0.939 & 0.158 & 0.886 \\ \midrule
    \texttt{all} & 0.915 & 0.119 & 0.844 \\
    \bottomrule
\end{tabular}}
\caption{FActscore F1 scores on reference factuality labels for claims in \textsc{ExpertQA}.}
\vspace{-10pt}
\label{tab:factuality}
\end{table}

In Table~\ref{tab:factuality}, we report the F1 scores of the factual (T) and non-factual (F) classes and the micro-averaged overall F1 scores of the FActScore factuality scores and the reference factuality labels. FActscore scores are thresholded at 0.5 to get binary scores and reference factuality labels are 1 if the claim's factuality is labeled as \textit{Probably correct} or \textit{Definitely correct}, and 0 otherwise.

We find that automatic factuality estimation struggles to identify non-factual claims. In particular, predicted labels have low recall of non-factual claims. This is more often the case for retrieve-and-read systems, where the answer is generated based on retrieved evidences. The other systems use GPT-4's parametric knowledge for answer generation, which could make it easier for a similar evaluator like ChatGPT to judge factuality.




\section{Long-form QA Evaluation}
\label{sec:lfqa}
A beneficial output of our annotation is the revised answers produced by annotators. These answers are verified to be factual and compose a new long-form QA dataset, \textsc{ExpertQA}.
We consider two splits for \textsc{ExpertQA} (both 80-10-10): a random split of the data and a domain-wise split, where 80\% of a field's data is included in the training set and 10\% is included in both validation and test sets.

\subsection{Evaluation Metrics}

For evaluation, we consider metrics based on similarity to a reference answer, i.e., ROUGE \cite{lin-2004-rouge} and those focused on evaluating factual consistency through QA pairs generated with a reference answer, i.e., QAFactEval \cite{fabbri-etal-2022-qafacteval}. 

\subsection{Baselines}

We finetune the following open-source language models: FlanT5-11B \cite{chung2022scaling},  Alpaca-7B \cite{alpaca}, Vicuna-7B \cite{vicuna2023} and LLaMa2-7B-Chat \cite{touvron2023llama}. We finetune these models with the same prompts as the ones used in their training (provided in Tables~\ref{tab:llama2_prompt},~\ref{tab:vicuna_prompt}). Further, we also report results with Llama2-70B-Chat without finetuning (marked *).

\begin{table}[!t]
\centering
\scalebox{0.8}{
\begin{tabular}{c|c|c|c|c|c}
    \textbf{Split} & \textbf{Model} & \textbf{R1} & \textbf{R2} & \textbf{RL} & \textbf{QFE} \\ \toprule
    \multirow{4}{*}{Random} & FlanT5-11B & 0.335 & 0.114 & 0.215 & 2.068 \\
                            & Vicuna-7B & 0.351 & 0.119 & 0.212 & 1.068 \\
                            & Llama2-7B & 0.362 & 0.125 & 0.219 & 1.985 \\
                            & Llama2-70B* & 0.320 & 0.101 & 0.181 & 1.050 \\
    \midrule
    \multirow{4}{*}{Domain} & FlanT5-11B & 0.324 & 0.107 & 0.210 & 1.538 \\
                            & Vicuna-7B & 0.359 & 0.120 & 0.213 & 1.739 \\
                            & Llama2-7B & 0.363 & 0.124 & 0.219 & 1.726 \\
                            & Llama2-70B* & 0.328 & 0.104 & 0.187 & 0.979 \\
    \bottomrule
\end{tabular}}
\caption{Long-form QA results (ROUGE scores and QAFactEval scores) after finetuning models on the random and domain splits of \textsc{ExpertQA}.}
\vspace{-10pt}
\label{tab:lfqa_results}
\end{table}

\subsection{Results}

Our results are shown in Table~\ref{tab:lfqa_results}. We find that both Llama2-7B and Vicuna-7B outperform FlanT5-11B despite the smaller model size, likely due to additional instruction finetuning for both those models. We observe that finetuning significantly improves performance (results without finetuning are in Table~\ref{tab:lfqa_results_full}), and Llama2-70B performs worse than finetuned systems under zero-shot prompting.

\section{Related Work}

\paragraph{Attribution Generation.} A few classes of systems have been proposed for generating attributions for model responses. This includes \textbf{vanilla LLM prompting} \cite{tay2022transformer}, where LLMs are prompted to return attributions with their answers, but the references are often hallucinated \cite{agrawal2023language}. On the other hand, \textbf{retrieve-and-read systems} \cite{guu2020retrieval,borgeaud2022improving,izacard2022few} first retrieve evidence relevant for a query, and generate an answer based on the retrieved evidence. These systems are sometimes trained on human demonstrations \cite{nakano2021webgpt, thoppilan2022lamda, menick2022teaching}. Finally, \textbf{post-hoc retrieval} \cite{gao2022rarr,he2022rethinking} involves retrieving attributions after answering a query. We consider all three classes of systems for sampling responses.

\paragraph{Attribution Analysis}

Prior work has conducted analysis of system-generated attributions \cite{rashkin2021measuring,bohnet2022attributed,dziri2022evaluating,chen2022propsegment,liu2023evaluating,muller2023evaluating,kamoi2023wice,kamalloo2023hagrid}.
These works suggest that systems are still far from providing precise attributions with sufficient recall for citeworthy statements. In our work, we recognize that this is problematic in specific domains where precision and recall are both critical.


\paragraph{Factuality Analysis.}

Factuality analysis of model generations has been conducted extensively in prior work \cite{thorne-etal-2018-fact,evans2021truthful,kryscinski-etal-2020-evaluating,maynez-etal-2020-faithfulness,pagnoni-etal-2021-understanding,lin2021truthfulqa,atanasova2022fact,muhlgay2023generating,tang2024tofueval,mishra2024finegrained}. \citet{peskoff-stewart-2023-credible} conduct a smaller-scale evaluation of modern LMs with 10 experts where they evaluate accuracy, among other qualities of answers.
The factuality labels collected as part of \textsc{ExpertQA} elicit a best-effort judgement of truthfulness of claims from domain experts. 
Prior work has also proposed methods to predict factuality of claims \cite{manakul2023selfcheckgpt, kadavath2022language, agrawal2023language, azaria2023internal, min2023fact, feng2023factkb, chen2023complex}. We use one such method \cite{min2023fact} to evaluate how well human labels in \textsc{ExpertQA} correlate with automatic judgements.
 
\paragraph{Long-form QA.} Existing long-form QA datasets are created using search queries \cite{nguyen2016ms, stelmakh2022asqa} and forums \cite{fan2019eli5}. Several issues have been identified with these datasets, such as vague questions and difficulty in verifying factual correctness \cite{krishna-etal-2021-hurdles}.
Keeping this in mind, we construct \textsc{ExpertQA} to cover practical information needs of experts along with fine-grained factuality judgements. \citet{xu2023critical} conduct expert evaluation of long-form answers and emphasize the importance of evaluating multiple aspects of answers, which are also considered in our work.

\paragraph{Domain-specific QA.} 

Several domain-specific QA datasets have been proposed, for domains such as medicine \cite{tsatsaronis2015overview,pampari-etal-2018-emrqa,jin-etal-2019-pubmedqa,jin2021disease,pmlr-v174-pal22a}, law \cite{guha2023legalbench},
technology \cite{dos2015learning} and others \cite{rogers2020getting,reddy-etal-2019-coqa,hendryckstest2021}. However, these datasets often have limited coverage of domains. \textsc{ExpertQA} contributes a unique combination of features by scaling the number of domains and providing attributions and factuality judgements.


\section{Conclusion and Future Work}
Our evaluation study suggests that although large language models show a lot of promise for aiding domain experts, there is large ground to cover in addressing the information needs of experts with factual and verifiable answers \cite{metzler2021rethinking}. Experts, on the other hand, should take responses from these systems with caution, because although attributed responses can seem trustworthy, the supporting references can often be inadequate to support claims.
We hope that our benchmark, \textsc{ExpertQA}, can benefit the community in building improved methods for attribution \& factuality estimation, and long-form question answering.

\section{Limitations}

\paragraph{Atomicity of Claims.} In most cases, claims in our dataset are sentences that may not represent singular information units. This lack of atomicity in claims means that properties such as factuality and attribution need to be judged exhaustively for a claim. Collecting human judgements for finer-grained atomic claims can be significantly more expensive and is not explored in this work.

\paragraph{Claim Extraction.} Extracting sentence-level claims from a generated answer for the purpose of evaluation is performed by using a sentence tokenizer. However, we note that existing tokenizers suffer from sentence tokenization errors (for example, when lists or tables are present in answers). This resulted in a small number of claims being excessively long and hard to evaluate.

\paragraph{Field Coverage.} Even though we tried to cover a wide range of fields in our dataset, we missed covering questions from certain fields. Finding experts from rarer fields can be especially hard. We will consider further expanding \textsc{ExpertQA} to more domains, so that it can be more broadly useful. In addition, the examples in our dataset represent the information needs of English-speaking annotators primarily based in Europe, the Americas and Africa.

\paragraph{Question Distribution.} We elicit questions from experts by asking them to formulate questions that have come up in their professional lives or questions they are genuinely curious about. This was aimed at modeling a more realistic information-seeking scenario through our annotation. However, it is not necessary that these questions would come from a natural distribution that would be found in query logs. Since having access to such data is not possible, we attempt to match the information-seeking scenario as closely as possible.

\paragraph{Subjectivity of labels.} Some of the properties of claims can elicit more subjective judgements, which can vary between experts from the same field. This subjectivity is not inherently captured in our data through multiple judgements, but we do estimate agreement using claims from engineering and medicine through our own labels (\S\ref{sec:manual}).



\section*{Acknowledgements}

First, we would like to thank the 484 annotators who took the time and effort out to help out with this study. This study would not have been possible without their contributions. We would also like to thank Artemis Panagopoulou, Alyssa Hwang, Nelson Liu and Chris Alberti for helpful comments and discussions.

\bibliography{anthology,custom}

\begin{thebibliography}{74}
\expandafter\ifx\csname natexlab\endcsname\relax\def\natexlab#1{#1}\fi

\bibitem[{Agrawal et~al.(2023)Agrawal, Mackey, and Kalai}]{agrawal2023language}
Ayush Agrawal, Lester Mackey, and Adam~Tauman Kalai. 2023.
\newblock \href {https://arxiv.org/abs/2305.18248} {Do language models know when they're hallucinating references?}
\newblock \emph{arXiv preprint arXiv:2305.18248}.

\bibitem[{Atanasova et~al.(2022)Atanasova, Simonsen, Lioma, and Augenstein}]{atanasova2022fact}
Pepa Atanasova, Jakob~Grue Simonsen, Christina Lioma, and Isabelle Augenstein. 2022.
\newblock \href {https://direct.mit.edu/tacl/article/doi/10.1162/tacl_a_00486/112498/Fact-Checking-with-Insufficient-Evidence} {Fact checking with insufficient evidence}.
\newblock \emph{Transactions of the Association for Computational Linguistics}, 10:746--763.

\bibitem[{Augenstein et~al.(2023)Augenstein, Baldwin, Cha, Chakraborty, Ciampaglia, Corney, DiResta, Ferrara, Hale, Halevy et~al.}]{augenstein2023factuality}
Isabelle Augenstein, Timothy Baldwin, Meeyoung Cha, Tanmoy Chakraborty, Giovanni~Luca Ciampaglia, David Corney, Renee DiResta, Emilio Ferrara, Scott Hale, Alon Halevy, et~al. 2023.
\newblock \href {https://arxiv.org/abs/2310.05189} {Factuality challenges in the era of large language models}.
\newblock \emph{arXiv preprint arXiv:2310.05189}.

\bibitem[{Azaria and Mitchell(2023)}]{azaria2023internal}
Amos Azaria and Tom Mitchell. 2023.
\newblock \href {https://arxiv.org/abs/2304.13734} {The internal state of an llm knows when its lying}.
\newblock \emph{arXiv preprint arXiv:2304.13734}.

\bibitem[{Birhane et~al.(2023)Birhane, Kasirzadeh, Leslie, and Wachter}]{birhane2023}
Abeba Birhane, Atoosa Kasirzadeh, David Leslie, and Sandra Wachter. 2023.
\newblock \href {https://doi.org/10.1038/s42254-023-00581-4} {Science in the age of large language models}.
\newblock \emph{Nature Reviews Physics}, 5(5):277--280.

\bibitem[{Bohnet et~al.(2022)Bohnet, Tran, Verga, Aharoni, Andor, Soares, Eisenstein, Ganchev, Herzig, Hui et~al.}]{bohnet2022attributed}
Bernd Bohnet, Vinh~Q Tran, Pat Verga, Roee Aharoni, Daniel Andor, Livio~Baldini Soares, Jacob Eisenstein, Kuzman Ganchev, Jonathan Herzig, Kai Hui, et~al. 2022.
\newblock \href {https://arxiv.org/abs/2212.08037} {Attributed question answering: Evaluation and modeling for attributed large language models}.
\newblock \emph{arXiv preprint arXiv:2212.08037}.

\bibitem[{Borgeaud et~al.(2022)Borgeaud, Mensch, Hoffmann, Cai, Rutherford, Millican, Van Den~Driessche, Lespiau, Damoc, Clark et~al.}]{borgeaud2022improving}
Sebastian Borgeaud, Arthur Mensch, Jordan Hoffmann, Trevor Cai, Eliza Rutherford, Katie Millican, George~Bm Van Den~Driessche, Jean-Baptiste Lespiau, Bogdan Damoc, Aidan Clark, et~al. 2022.
\newblock \href {https://proceedings.mlr.press/v162/borgeaud22a.html} {Improving language models by retrieving from trillions of tokens}.
\newblock In \emph{International conference on machine learning}, pages 2206--2240. PMLR.

\bibitem[{Brown et~al.(2020)Brown, Mann, Ryder, Subbiah, Kaplan, Dhariwal, Neelakantan, Shyam, Sastry, Askell et~al.}]{brown2020language}
Tom Brown, Benjamin Mann, Nick Ryder, Melanie Subbiah, Jared~D Kaplan, Prafulla Dhariwal, Arvind Neelakantan, Pranav Shyam, Girish Sastry, Amanda Askell, et~al. 2020.
\newblock \href {https://proceedings.neurips.cc/paper/2020/hash/1457c0d6bfcb4967418bfb8ac142f64a-Abstract.html} {Language models are few-shot learners}.
\newblock \emph{Advances in neural information processing systems}, 33:1877--1901.

\bibitem[{Chen et~al.(2017)Chen, Fisch, Weston, and Bordes}]{chen-etal-2017-reading}
Danqi Chen, Adam Fisch, Jason Weston, and Antoine Bordes. 2017.
\newblock \href {https://doi.org/10.18653/v1/P17-1171} {Reading {W}ikipedia to answer open-domain questions}.
\newblock In \emph{Proceedings of the 55th Annual Meeting of the Association for Computational Linguistics (Volume 1: Long Papers)}, pages 1870--1879, Vancouver, Canada. Association for Computational Linguistics.

\bibitem[{Chen et~al.(2023)Chen, Kim, Sriram, Durrett, and Choi}]{chen2023complex}
Jifan Chen, Grace Kim, Aniruddh Sriram, Greg Durrett, and Eunsol Choi. 2023.
\newblock \href {https://arxiv.org/abs/2305.11859} {Complex claim verification with evidence retrieved in the wild}.
\newblock \emph{arXiv preprint arXiv:2305.11859}.

\bibitem[{Chen et~al.(2022)Chen, Buthpitiya, Fabrikant, Roth, and Schuster}]{chen2022propsegment}
Sihao Chen, Senaka Buthpitiya, Alex Fabrikant, Dan Roth, and Tal Schuster. 2022.
\newblock \href {https://arxiv.org/abs/2212.10750} {Propsegment: A large-scale corpus for proposition-level segmentation and entailment recognition}.
\newblock \emph{arXiv preprint arXiv:2212.10750}.

\bibitem[{Chiang et~al.(2023)Chiang, Li, Lin, Sheng, Wu, Zhang, Zheng, Zhuang, Zhuang, Gonzalez, Stoica, and Xing}]{vicuna2023}
Wei-Lin Chiang, Zhuohan Li, Zi~Lin, Ying Sheng, Zhanghao Wu, Hao Zhang, Lianmin Zheng, Siyuan Zhuang, Yonghao Zhuang, Joseph~E. Gonzalez, Ion Stoica, and Eric~P. Xing. 2023.
\newblock \href {https://lmsys.org/blog/2023-03-30-vicuna/} {Vicuna: An open-source chatbot impressing gpt-4 with 90\%* chatgpt quality}.

\bibitem[{Chung et~al.(2022)Chung, Hou, Longpre, Zoph, Tay, Fedus, Li, Wang, Dehghani, Brahma et~al.}]{chung2022scaling}
Hyung~Won Chung, Le~Hou, Shayne Longpre, Barret Zoph, Yi~Tay, William Fedus, Eric Li, Xuezhi Wang, Mostafa Dehghani, Siddhartha Brahma, et~al. 2022.
\newblock \href {https://arxiv.org/abs/2210.11416} {Scaling instruction-finetuned language models}.
\newblock \emph{arXiv preprint arXiv:2210.11416}.

\bibitem[{Dash et~al.(2023)Dash, Thapa, Banda, Swaminathan, Cheatham, Kashyap, Kotecha, Chen, Gombar, Downing et~al.}]{dash2023evaluation}
Debadutta Dash, Rahul Thapa, Juan~M Banda, Akshay Swaminathan, Morgan Cheatham, Mehr Kashyap, Nikesh Kotecha, Jonathan~H Chen, Saurabh Gombar, Lance Downing, et~al. 2023.
\newblock \href {https://arxiv.org/abs/2304.13714} {Evaluation of gpt-3.5 and gpt-4 for supporting real-world information needs in healthcare delivery}.
\newblock \emph{arXiv preprint arXiv:2304.13714}.

\bibitem[{Dos~Santos et~al.(2015)Dos~Santos, Barbosa, Bogdanova, and Zadrozny}]{dos2015learning}
Cicero Dos~Santos, Luciano Barbosa, Dasha Bogdanova, and Bianca Zadrozny. 2015.
\newblock \href {https://aclanthology.org/P15-2114/} {Learning hybrid representations to retrieve semantically equivalent questions}.
\newblock In \emph{Proceedings of the 53rd Annual Meeting of the Association for Computational Linguistics and the 7th International Joint Conference on Natural Language Processing (Volume 2: Short Papers)}, pages 694--699.

\bibitem[{Dziri et~al.(2022)Dziri, Rashkin, Linzen, and Reitter}]{dziri2022evaluating}
Nouha Dziri, Hannah Rashkin, Tal Linzen, and David Reitter. 2022.
\newblock \href {https://direct.mit.edu/tacl/article/doi/10.1162/tacl_a_00506/113023/Evaluating-Attribution-in-Dialogue-Systems-The} {Evaluating attribution in dialogue systems: The begin benchmark}.
\newblock \emph{Transactions of the Association for Computational Linguistics}, 10:1066--1083.

\bibitem[{Evans et~al.(2021)Evans, Cotton-Barratt, Finnveden, Bales, Balwit, Wills, Righetti, and Saunders}]{evans2021truthful}
Owain Evans, Owen Cotton-Barratt, Lukas Finnveden, Adam Bales, Avital Balwit, Peter Wills, Luca Righetti, and William Saunders. 2021.
\newblock \href {https://arxiv.org/abs/2110.06674} {Truthful ai: Developing and governing ai that does not lie}.
\newblock \emph{arXiv preprint arXiv:2110.06674}.

\bibitem[{Fabbri et~al.(2022)Fabbri, Wu, Liu, and Xiong}]{fabbri-etal-2022-qafacteval}
Alexander~R. Fabbri, Chien-Sheng Wu, Wenhao Liu, and Caiming Xiong. 2022.
\newblock \href {http://arxiv.org/abs/2112.08542} {Qafacteval: Improved qa-based factual consistency evaluation for summarization}.

\bibitem[{Fan et~al.(2019)Fan, Jernite, Perez, Grangier, Weston, and Auli}]{fan2019eli5}
Angela Fan, Yacine Jernite, Ethan Perez, David Grangier, Jason Weston, and Michael Auli. 2019.
\newblock \href {https://arxiv.org/abs/1907.09190} {Eli5: Long form question answering}.
\newblock \emph{arXiv preprint arXiv:1907.09190}.

\bibitem[{Feng et~al.(2023)Feng, Balachandran, Bai, and Tsvetkov}]{feng2023factkb}
Shangbin Feng, Vidhisha Balachandran, Yuyang Bai, and Yulia Tsvetkov. 2023.
\newblock \href {https://arxiv.org/abs/2305.08281} {Factkb: Generalizable factuality evaluation using language models enhanced with factual knowledge}.
\newblock \emph{arXiv preprint arXiv:2305.08281}.

\bibitem[{Gao et~al.(2023)Gao, Dai, Pasupat, Chen, Chaganty, Fan, Zhao, Lao, Lee, Juan, and Guu}]{gao2022rarr}
Luyu Gao, Zhuyun Dai, Panupong Pasupat, Anthony Chen, Arun~Tejasvi Chaganty, Yicheng Fan, Vincent Zhao, Ni~Lao, Hongrae Lee, Da-Cheng Juan, and Kelvin Guu. 2023.
\newblock \href {https://doi.org/10.18653/v1/2023.acl-long.910} {{RARR}: Researching and revising what language models say, using language models}.
\newblock In \emph{Proceedings of the 61st Annual Meeting of the Association for Computational Linguistics (Volume 1: Long Papers)}, pages 16477--16508, Toronto, Canada. Association for Computational Linguistics.

\bibitem[{Guha et~al.(2023)Guha, Nyarko, Ho, R{\'e}, Chilton, Narayana, Chohlas-Wood, Peters, Waldon, Rockmore et~al.}]{guha2023legalbench}
Neel Guha, Julian Nyarko, Daniel~E Ho, Christopher R{\'e}, Adam Chilton, Aditya Narayana, Alex Chohlas-Wood, Austin Peters, Brandon Waldon, Daniel~N Rockmore, et~al. 2023.
\newblock \href {https://arxiv.org/abs/2308.11462} {Legalbench: A collaboratively built benchmark for measuring legal reasoning in large language models}.
\newblock \emph{arXiv preprint arXiv:2308.11462}.

\bibitem[{Guu et~al.(2020)Guu, Lee, Tung, Pasupat, and Chang}]{guu2020retrieval}
Kelvin Guu, Kenton Lee, Zora Tung, Panupong Pasupat, and Mingwei Chang. 2020.
\newblock \href {http://proceedings.mlr.press/v119/guu20a.html} {Retrieval augmented language model pre-training}.
\newblock In \emph{International conference on machine learning}, pages 3929--3938. PMLR.

\bibitem[{He et~al.(2022)He, Zhang, and Roth}]{he2022rethinking}
Hangfeng He, Hongming Zhang, and Dan Roth. 2022.
\newblock \href {https://arxiv.org/abs/2301.00303} {Rethinking with retrieval: Faithful large language model inference}.
\newblock \emph{arXiv preprint arXiv:2301.00303}.

\bibitem[{Hendrycks et~al.(2021)Hendrycks, Burns, Basart, Zou, Mazeika, Song, and Steinhardt}]{hendryckstest2021}
Dan Hendrycks, Collin Burns, Steven Basart, Andy Zou, Mantas Mazeika, Dawn Song, and Jacob Steinhardt. 2021.
\newblock \href {https://arxiv.org/abs/2009.03300} {Measuring massive multitask language understanding}.
\newblock \emph{Proceedings of the International Conference on Learning Representations (ICLR)}.

\bibitem[{Honnibal and Montani(2017)}]{spacy2}
Matthew Honnibal and Ines Montani. 2017.
\newblock \href {https://spacy.io/} {{spaCy 2}: Natural language understanding with {B}loom embeddings, convolutional neural networks and incremental parsing}.
\newblock To appear.

\bibitem[{Honovich et~al.(2022)Honovich, Aharoni, Herzig, Taitelbaum, Kukliansy, Cohen, Scialom, Szpektor, Hassidim, and Matias}]{honovich-etal-2022-true-evaluating}
Or~Honovich, Roee Aharoni, Jonathan Herzig, Hagai Taitelbaum, Doron Kukliansy, Vered Cohen, Thomas Scialom, Idan Szpektor, Avinatan Hassidim, and Yossi Matias. 2022.
\newblock \href {https://doi.org/10.18653/v1/2022.naacl-main.287} {{TRUE}: Re-evaluating factual consistency evaluation}.
\newblock In \emph{Proceedings of the 2022 Conference of the North American Chapter of the Association for Computational Linguistics: Human Language Technologies}, pages 3905--3920, Seattle, United States. Association for Computational Linguistics.

\bibitem[{Izacard et~al.(2022)Izacard, Lewis, Lomeli, Hosseini, Petroni, Schick, Dwivedi-Yu, Joulin, Riedel, and Grave}]{izacard2022few}
Gautier Izacard, Patrick Lewis, Maria Lomeli, Lucas Hosseini, Fabio Petroni, Timo Schick, Jane Dwivedi-Yu, Armand Joulin, Sebastian Riedel, and Edouard Grave. 2022.
\newblock \href {https://arxiv.org/abs/2208.03299} {Few-shot learning with retrieval augmented language models}.
\newblock \emph{arXiv preprint arXiv:2208.03299}.

\bibitem[{Jin et~al.(2021)Jin, Pan, Oufattole, Weng, Fang, and Szolovits}]{jin2021disease}
Di~Jin, Eileen Pan, Nassim Oufattole, Wei-Hung Weng, Hanyi Fang, and Peter Szolovits. 2021.
\newblock \href {https://www.mdpi.com/2076-3417/11/14/6421} {What disease does this patient have? a large-scale open domain question answering dataset from medical exams}.
\newblock \emph{Applied Sciences}, 11(14):6421.

\bibitem[{Jin et~al.(2019)Jin, Dhingra, Liu, Cohen, and Lu}]{jin-etal-2019-pubmedqa}
Qiao Jin, Bhuwan Dhingra, Zhengping Liu, William Cohen, and Xinghua Lu. 2019.
\newblock \href {https://doi.org/10.18653/v1/D19-1259} {{P}ub{M}ed{QA}: A dataset for biomedical research question answering}.
\newblock In \emph{Proceedings of the 2019 Conference on Empirical Methods in Natural Language Processing and the 9th International Joint Conference on Natural Language Processing (EMNLP-IJCNLP)}, pages 2567--2577, Hong Kong, China. Association for Computational Linguistics.

\bibitem[{Kadavath et~al.(2022)Kadavath, Conerly, Askell, Henighan, Drain, Perez, Schiefer, Dodds, DasSarma, Tran-Johnson et~al.}]{kadavath2022language}
Saurav Kadavath, Tom Conerly, Amanda Askell, Tom Henighan, Dawn Drain, Ethan Perez, Nicholas Schiefer, Zac~Hatfield Dodds, Nova DasSarma, Eli Tran-Johnson, et~al. 2022.
\newblock \href {https://arxiv.org/abs/2207.05221} {Language models (mostly) know what they know}.
\newblock \emph{arXiv preprint arXiv:2207.05221}.

\bibitem[{Kamalloo et~al.(2023)Kamalloo, Jafari, Zhang, Thakur, and Lin}]{kamalloo2023hagrid}
Ehsan Kamalloo, Aref Jafari, Xinyu Zhang, Nandan Thakur, and Jimmy Lin. 2023.
\newblock \href {https://arxiv.org/abs/2307.16883} {Hagrid: A human-llm collaborative dataset for generative information-seeking with attribution}.
\newblock \emph{arXiv preprint arXiv:2307.16883}.

\bibitem[{Kamoi et~al.(2023)Kamoi, Goyal, Rodriguez, and Durrett}]{kamoi2023wice}
Ryo Kamoi, Tanya Goyal, Juan~Diego Rodriguez, and Greg Durrett. 2023.
\newblock \href {https://arxiv.org/pdf/2303.01432.pdf} {Wice: Real-world entailment for claims in wikipedia}.
\newblock \emph{arXiv preprint arXiv:2303.01432}.

\bibitem[{Krenn et~al.(2022)Krenn, Pollice, Guo, Aldeghi, Cervera-Lierta, Friederich, dos Passos Gomes, H{\"a}se, Jinich, Nigam, Yao, and Aspuru-Guzik}]{krenn2022}
Mario Krenn, Robert Pollice, Si~Yue Guo, Matteo Aldeghi, Alba Cervera-Lierta, Pascal Friederich, Gabriel dos Passos Gomes, Florian H{\"a}se, Adrian Jinich, AkshatKumar Nigam, Zhenpeng Yao, and Al{\'a}n Aspuru-Guzik. 2022.
\newblock \href {https://doi.org/10.1038/s42254-022-00518-3} {On scientific understanding with artificial intelligence}.
\newblock \emph{Nature Reviews Physics}, 4(12):761--769.

\bibitem[{Krishna et~al.(2021)Krishna, Roy, and Iyyer}]{krishna-etal-2021-hurdles}
Kalpesh Krishna, Aurko Roy, and Mohit Iyyer. 2021.
\newblock \href {https://doi.org/10.18653/v1/2021.naacl-main.393} {Hurdles to progress in long-form question answering}.
\newblock In \emph{Proceedings of the 2021 Conference of the North American Chapter of the Association for Computational Linguistics: Human Language Technologies}, pages 4940--4957, Online. Association for Computational Linguistics.

\bibitem[{Kryscinski et~al.(2020)Kryscinski, McCann, Xiong, and Socher}]{kryscinski-etal-2020-evaluating}
Wojciech Kryscinski, Bryan McCann, Caiming Xiong, and Richard Socher. 2020.
\newblock \href {https://doi.org/10.18653/v1/2020.emnlp-main.750} {Evaluating the factual consistency of abstractive text summarization}.
\newblock In \emph{Proceedings of the 2020 Conference on Empirical Methods in Natural Language Processing (EMNLP)}, pages 9332--9346, Online. Association for Computational Linguistics.

\bibitem[{Lee et~al.(2023)Lee, Bubeck, and Petro}]{lee2023benefits}
Peter Lee, Sebastien Bubeck, and Joseph Petro. 2023.
\newblock \href {https://www.nejm.org/doi/full/10.1056/NEJMsr2214184} {Benefits, limits, and risks of gpt-4 as an ai chatbot for medicine}.
\newblock \emph{New England Journal of Medicine}, 388(13):1233--1239.

\bibitem[{Lin(2004)}]{lin-2004-rouge}
Chin-Yew Lin. 2004.
\newblock \href {https://aclanthology.org/W04-1013} {{ROUGE}: A package for automatic evaluation of summaries}.
\newblock In \emph{Text Summarization Branches Out}, pages 74--81, Barcelona, Spain. Association for Computational Linguistics.

\bibitem[{Lin et~al.(2021)Lin, Hilton, and Evans}]{lin2021truthfulqa}
Stephanie Lin, Jacob Hilton, and Owain Evans. 2021.
\newblock \href {https://arxiv.org/abs/2109.07958} {Truthfulqa: Measuring how models mimic human falsehoods}.
\newblock \emph{arXiv preprint arXiv:2109.07958}.

\bibitem[{Liu et~al.(2023)Liu, Zhang, and Liang}]{liu2023evaluating}
Nelson~F Liu, Tianyi Zhang, and Percy Liang. 2023.
\newblock \href {https://arxiv.org/abs/2304.09848} {Evaluating verifiability in generative search engines}.
\newblock \emph{arXiv preprint arXiv:2304.09848}.

\bibitem[{Manakul et~al.(2023)Manakul, Liusie, and Gales}]{manakul2023selfcheckgpt}
Potsawee Manakul, Adian Liusie, and Mark~JF Gales. 2023.
\newblock \href {https://arxiv.org/abs/2303.08896} {Selfcheckgpt: Zero-resource black-box hallucination detection for generative large language models}.
\newblock \emph{arXiv preprint arXiv:2303.08896}.

\bibitem[{Maynez et~al.(2020)Maynez, Narayan, Bohnet, and McDonald}]{maynez-etal-2020-faithfulness}
Joshua Maynez, Shashi Narayan, Bernd Bohnet, and Ryan McDonald. 2020.
\newblock \href {https://doi.org/10.18653/v1/2020.acl-main.173} {On faithfulness and factuality in abstractive summarization}.
\newblock In \emph{Proceedings of the 58th Annual Meeting of the Association for Computational Linguistics}, pages 1906--1919, Online. Association for Computational Linguistics.

\bibitem[{Menick et~al.(2022)Menick, Trebacz, Mikulik, Aslanides, Song, Chadwick, Glaese, Young, Campbell-Gillingham, Irving et~al.}]{menick2022teaching}
Jacob Menick, Maja Trebacz, Vladimir Mikulik, John Aslanides, Francis Song, Martin Chadwick, Mia Glaese, Susannah Young, Lucy Campbell-Gillingham, Geoffrey Irving, et~al. 2022.
\newblock \href {https://arxiv.org/abs/2203.11147} {Teaching language models to support answers with verified quotes}.
\newblock \emph{arXiv preprint arXiv:2203.11147}.

\bibitem[{Metzler et~al.(2021)Metzler, Tay, Bahri, and Najork}]{metzler2021rethinking}
Donald Metzler, Yi~Tay, Dara Bahri, and Marc Najork. 2021.
\newblock \href {https://arxiv.org/abs/2105.02274} {Rethinking search: making domain experts out of dilettantes}.
\newblock In \emph{Acm sigir forum}, volume~55, pages 1--27. ACM New York, NY, USA.

\bibitem[{Min et~al.(2023)Min, Krishna, Lyu, Lewis, Yih, Koh, Iyyer, Zettlemoyer, and Hajishirzi}]{min2023fact}
Sewon Min, Kalpesh Krishna, Xinxi Lyu, Mike Lewis, Wen-tau Yih, Pang~Wei Koh, Mohit Iyyer, Luke Zettlemoyer, and Hannaneh Hajishirzi. 2023.
\newblock \href {https://arxiv.org/abs/2305.14251v1} {Factscore: Fine-grained atomic evaluation of factual precision in long form text generation}.
\newblock \emph{arXiv preprint arXiv:2305.14251v1}.

\bibitem[{Mishra et~al.(2024)Mishra, Asai, Balachandran, Wang, Neubig, Tsvetkov, and Hajishirzi}]{mishra2024finegrained}
Abhika Mishra, Akari Asai, Vidhisha Balachandran, Yizhong Wang, Graham Neubig, Yulia Tsvetkov, and Hannaneh Hajishirzi. 2024.
\newblock \href {https://arxiv.org/abs/2401.06855} {Fine-grained hallucinations detections}.
\newblock \emph{arXiv preprint}.

\bibitem[{Muhlgay et~al.(2023)Muhlgay, Ram, Magar, Levine, Ratner, Belinkov, Abend, Leyton-Brown, Shashua, and Shoham}]{muhlgay2023generating}
Dor Muhlgay, Ori Ram, Inbal Magar, Yoav Levine, Nir Ratner, Yonatan Belinkov, Omri Abend, Kevin Leyton-Brown, Amnon Shashua, and Yoav Shoham. 2023.
\newblock \href {https://arxiv.org/abs/2307.06908} {Generating benchmarks for factuality evaluation of language models}.
\newblock \emph{arXiv preprint arXiv:2307.06908}.

\bibitem[{Muller et~al.(2023)Muller, Wieting, Clark, Kwiatkowski, Ruder, Soares, Aharoni, Herzig, and Wang}]{muller2023evaluating}
Benjamin Muller, John Wieting, Jonathan~H Clark, Tom Kwiatkowski, Sebastian Ruder, Livio~Baldini Soares, Roee Aharoni, Jonathan Herzig, and Xinyi Wang. 2023.
\newblock \href {https://arxiv.org/abs/2305.14332} {Evaluating and modeling attribution for cross-lingual question answering}.
\newblock \emph{arXiv preprint arXiv:2305.14332}.

\bibitem[{Nakano et~al.(2021)Nakano, Hilton, Balaji, Wu, Ouyang, Kim, Hesse, Jain, Kosaraju, Saunders et~al.}]{nakano2021webgpt}
Reiichiro Nakano, Jacob Hilton, Suchir Balaji, Jeff Wu, Long Ouyang, Christina Kim, Christopher Hesse, Shantanu Jain, Vineet Kosaraju, William Saunders, et~al. 2021.
\newblock \href {https://arxiv.org/abs/2112.09332} {Webgpt: Browser-assisted question-answering with human feedback}.
\newblock \emph{arXiv preprint arXiv:2112.09332}.

\bibitem[{Nguyen et~al.(2016)Nguyen, Rosenberg, Song, Gao, Tiwary, Majumder, and Deng}]{nguyen2016ms}
Tri Nguyen, Mir Rosenberg, Xia Song, Jianfeng Gao, Saurabh Tiwary, Rangan Majumder, and Li~Deng. 2016.
\newblock \href {https://ceur-ws.org/Vol-1773/CoCoNIPS_2016_paper9.pdf} {Ms marco: A human generated machine reading comprehension dataset}.
\newblock \emph{choice}, 2640:660.

\bibitem[{OpenAI(2023)}]{OpenAI2023GPT4TR}
OpenAI. 2023.
\newblock \href {https://arxiv.org/abs/2303.08774} {Gpt-4 technical report}.
\newblock \emph{ArXiv}, abs/2303.08774.

\bibitem[{Owens(2023)}]{owens2023nature}
Brian Owens. 2023.
\newblock \href {https://www.nature.com/articles/d41586-023-00500-8} {How nature readers are using chatgpt}.
\newblock \emph{Nature}.

\bibitem[{Pagnoni et~al.(2021)Pagnoni, Balachandran, and Tsvetkov}]{pagnoni-etal-2021-understanding}
Artidoro Pagnoni, Vidhisha Balachandran, and Yulia Tsvetkov. 2021.
\newblock \href {https://doi.org/10.18653/v1/2021.naacl-main.383} {Understanding factuality in abstractive summarization with {FRANK}: A benchmark for factuality metrics}.
\newblock In \emph{Proceedings of the 2021 Conference of the North American Chapter of the Association for Computational Linguistics: Human Language Technologies}, pages 4812--4829, Online. Association for Computational Linguistics.

\bibitem[{Pal et~al.(2022)Pal, Umapathi, and Sankarasubbu}]{pmlr-v174-pal22a}
Ankit Pal, Logesh~Kumar Umapathi, and Malaikannan Sankarasubbu. 2022.
\newblock \href {https://proceedings.mlr.press/v174/pal22a.html} {Medmcqa: A large-scale multi-subject multi-choice dataset for medical domain question answering}.
\newblock In \emph{Proceedings of the Conference on Health, Inference, and Learning}, volume 174 of \emph{Proceedings of Machine Learning Research}, pages 248--260. PMLR.

\bibitem[{Pampari et~al.(2018)Pampari, Raghavan, Liang, and Peng}]{pampari-etal-2018-emrqa}
Anusri Pampari, Preethi Raghavan, Jennifer Liang, and Jian Peng. 2018.
\newblock \href {https://doi.org/10.18653/v1/D18-1258} {emr{QA}: A large corpus for question answering on electronic medical records}.
\newblock In \emph{Proceedings of the 2018 Conference on Empirical Methods in Natural Language Processing}, pages 2357--2368, Brussels, Belgium. Association for Computational Linguistics.

\bibitem[{Peskoff and Stewart(2023)}]{peskoff-stewart-2023-credible}
Denis Peskoff and Brandon Stewart. 2023.
\newblock \href {https://doi.org/10.18653/v1/2023.acl-short.37} {Credible without credit: Domain experts assess generative language models}.
\newblock In \emph{Proceedings of the 61st Annual Meeting of the Association for Computational Linguistics (Volume 2: Short Papers)}, pages 427--438, Toronto, Canada. Association for Computational Linguistics.

\bibitem[{Piktus et~al.(2021)Piktus, Petroni, Karpukhin, Okhonko, Broscheit, Izacard, Lewis, Oguz, Grave, Yih, and Riedel}]{sphere}
Aleksandra Piktus, Fabio Petroni, Vladimir Karpukhin, Dmytro Okhonko, Samuel Broscheit, Gautier Izacard, Patrick Lewis, Barlas Oguz, Edouard Grave, Wen{-}tau Yih, and Sebastian Riedel. 2021.
\newblock \href {http://arxiv.org/abs/2112.09924} {The web is your oyster - knowledge-intensive {NLP} against a very large web corpus}.
\newblock \emph{CoRR}, abs/2112.09924.

\bibitem[{Rajbhandari et~al.(2020)Rajbhandari, Rasley, Ruwase, and He}]{rajbhandari2020zero}
Samyam Rajbhandari, Jeff Rasley, Olatunji Ruwase, and Yuxiong He. 2020.
\newblock \href {https://ieeexplore.ieee.org/abstract/document/9355301/} {Zero: Memory optimizations toward training trillion parameter models}.
\newblock In \emph{SC20: International Conference for High Performance Computing, Networking, Storage and Analysis}, pages 1--16. IEEE.

\bibitem[{Rashkin et~al.(2021)Rashkin, Nikolaev, Lamm, Aroyo, Collins, Das, Petrov, Tomar, Turc, and Reitter}]{rashkin2021measuring}
Hannah Rashkin, Vitaly Nikolaev, Matthew Lamm, Lora Aroyo, Michael Collins, Dipanjan Das, Slav Petrov, Gaurav~Singh Tomar, Iulia Turc, and David Reitter. 2021.
\newblock \href {https://arxiv.org/abs/2112.12870} {Measuring attribution in natural language generation models}.
\newblock \emph{arXiv preprint arXiv:2112.12870}.

\bibitem[{Reddy et~al.(2019)Reddy, Chen, and Manning}]{reddy-etal-2019-coqa}
Siva Reddy, Danqi Chen, and Christopher~D. Manning. 2019.
\newblock \href {https://doi.org/10.1162/tacl_a_00266} {{C}o{QA}: A conversational question answering challenge}.
\newblock \emph{Transactions of the Association for Computational Linguistics}, 7:249--266.

\bibitem[{Robertson et~al.(2009)Robertson, Zaragoza et~al.}]{robertson2009probabilistic}
Stephen Robertson, Hugo Zaragoza, et~al. 2009.
\newblock \href {https://www.nowpublishers.com/article/Details/INR-019} {The probabilistic relevance framework: Bm25 and beyond}.
\newblock \emph{Foundations and Trends{\textregistered} in Information Retrieval}, 3(4):333--389.

\bibitem[{Rogers et~al.(2020)Rogers, Kovaleva, Downey, and Rumshisky}]{rogers2020getting}
Anna Rogers, Olga Kovaleva, Matthew Downey, and Anna Rumshisky. 2020.
\newblock \href {https://ojs.aaai.org/index.php/AAAI/article/view/6398/6254} {Getting closer to ai complete question answering: A set of prerequisite real tasks}.
\newblock In \emph{Proceedings of the AAAI conference on artificial intelligence}, volume~34, pages 8722--8731.

\bibitem[{Rose and Levinson(2004)}]{rose2004understanding}
Daniel~E Rose and Danny Levinson. 2004.
\newblock \href {https://dl.acm.org/doi/abs/10.1145/988672.988675} {Understanding user goals in web search}.
\newblock In \emph{Proceedings of the 13th international conference on World Wide Web}, pages 13--19.

\bibitem[{Schuster et~al.(2022)Schuster, Chen, Buthpitiya, Fabrikant, and Metzler}]{schuster-etal-2022-stretching}
Tal Schuster, Sihao Chen, Senaka Buthpitiya, Alex Fabrikant, and Donald Metzler. 2022.
\newblock \href {https://aclanthology.org/2022.findings-emnlp.28} {Stretching sentence-pair {NLI} models to reason over long documents and clusters}.
\newblock In \emph{Findings of the Association for Computational Linguistics: EMNLP 2022}, pages 394--412, Abu Dhabi, United Arab Emirates. Association for Computational Linguistics.

\bibitem[{Stelmakh et~al.(2022)Stelmakh, Luan, Dhingra, and Chang}]{stelmakh2022asqa}
Ivan Stelmakh, Yi~Luan, Bhuwan Dhingra, and Ming-Wei Chang. 2022.
\newblock \href {https://arxiv.org/abs/2204.06092} {Asqa: Factoid questions meet long-form answers}.
\newblock \emph{arXiv preprint arXiv:2204.06092}.

\bibitem[{Tang et~al.(2024)Tang, Shalyminov, mei Wong, Burnsky, Vincent, Yang, Singh, Feng, Song, Su, Sun, Zhang, Mansour, and McKeown}]{tang2024tofueval}
Liyan Tang, Igor Shalyminov, Amy~Wing mei Wong, Jon Burnsky, Jake~W. Vincent, Yu'an Yang, Siffi Singh, Song Feng, Hwanjun Song, Hang Su, Lijia Sun, Yi~Zhang, Saab Mansour, and Kathleen McKeown. 2024.
\newblock \href {http://arxiv.org/abs/2402.13249} {Tofueval: Evaluating hallucinations of llms on topic-focused dialogue summarization}.

\bibitem[{Taori et~al.(2023)Taori, Gulrajani, Zhang, Dubois, Li, Guestrin, Liang, and Hashimoto}]{alpaca}
Rohan Taori, Ishaan Gulrajani, Tianyi Zhang, Yann Dubois, Xuechen Li, Carlos Guestrin, Percy Liang, and Tatsunori~B. Hashimoto. 2023.
\newblock Stanford alpaca: An instruction-following llama model.
\newblock \url{https://github.com/tatsu-lab/stanford_alpaca}.

\bibitem[{Tay et~al.(2022)Tay, Tran, Dehghani, Ni, Bahri, Mehta, Qin, Hui, Zhao, Gupta et~al.}]{tay2022transformer}
Yi~Tay, Vinh Tran, Mostafa Dehghani, Jianmo Ni, Dara Bahri, Harsh Mehta, Zhen Qin, Kai Hui, Zhe Zhao, Jai Gupta, et~al. 2022.
\newblock \href {https://proceedings.neurips.cc/paper_files/paper/2022/hash/892840a6123b5ec99ebaab8be1530fba-Abstract-Conference.html} {Transformer memory as a differentiable search index}.
\newblock \emph{Advances in Neural Information Processing Systems}, 35:21831--21843.

\bibitem[{Thoppilan et~al.(2022)Thoppilan, De~Freitas, Hall, Shazeer, Kulshreshtha, Cheng, Jin, Bos, Baker, Du et~al.}]{thoppilan2022lamda}
Romal Thoppilan, Daniel De~Freitas, Jamie Hall, Noam Shazeer, Apoorv Kulshreshtha, Heng-Tze Cheng, Alicia Jin, Taylor Bos, Leslie Baker, Yu~Du, et~al. 2022.
\newblock \href {https://arxiv.org/abs/2201.08239} {Lamda: Language models for dialog applications}.
\newblock \emph{arXiv preprint arXiv:2201.08239}.

\bibitem[{Thorne et~al.(2018)Thorne, Vlachos, Cocarascu, Christodoulopoulos, and Mittal}]{thorne-etal-2018-fact}
James Thorne, Andreas Vlachos, Oana Cocarascu, Christos Christodoulopoulos, and Arpit Mittal. 2018.
\newblock \href {https://doi.org/10.18653/v1/W18-5501} {The fact extraction and {VER}ification ({FEVER}) shared task}.
\newblock In \emph{Proceedings of the First Workshop on Fact Extraction and {VER}ification ({FEVER})}, pages 1--9, Brussels, Belgium. Association for Computational Linguistics.

\bibitem[{Touvron et~al.(2023)Touvron, Martin, Stone, Albert, Almahairi, Babaei, Bashlykov, Batra, Bhargava, Bhosale et~al.}]{touvron2023llama}
Hugo Touvron, Louis Martin, Kevin Stone, Peter Albert, Amjad Almahairi, Yasmine Babaei, Nikolay Bashlykov, Soumya Batra, Prajjwal Bhargava, Shruti Bhosale, et~al. 2023.
\newblock \href {https://arxiv.org/abs/2307.09288} {Llama 2: Open foundation and fine-tuned chat models}.
\newblock \emph{arXiv preprint arXiv:2307.09288}.

\bibitem[{Tsatsaronis et~al.(2015)Tsatsaronis, Balikas, Malakasiotis, Partalas, Zschunke, Alvers, Weissenborn, Krithara, Petridis, Polychronopoulos et~al.}]{tsatsaronis2015overview}
George Tsatsaronis, Georgios Balikas, Prodromos Malakasiotis, Ioannis Partalas, Matthias Zschunke, Michael~R Alvers, Dirk Weissenborn, Anastasia Krithara, Sergios Petridis, Dimitris Polychronopoulos, et~al. 2015.
\newblock \href {https://bmcbioinformatics.biomedcentral.com/articles/10.1186/s12859-015-0564-6} {An overview of the bioasq large-scale biomedical semantic indexing and question answering competition}.
\newblock \emph{BMC bioinformatics}, 16(1):1--28.

\bibitem[{Volokh(2023)}]{volokh2023libel}
Eugene Volokh. 2023.
\newblock \href {https://www2.law.ucla.edu/volokh/ailibel.pdf} {Large libel models? liability for ai output}.

\bibitem[{Xu et~al.(2023)Xu, Song, Iyyer, and Choi}]{xu2023critical}
Fangyuan Xu, Yixiao Song, Mohit Iyyer, and Eunsol Choi. 2023.
\newblock \href {https://aclanthology.org/2023.acl-long.181} {A critical evaluation of evaluations for long-form question answering}.
\newblock In \emph{Proceedings of the 61st Annual Meeting of the Association for Computational Linguistics (Volume 1: Long Papers)}, pages 3225--3245, Toronto, Canada. Association for Computational Linguistics.

\end{thebibliography}
\bibliographystyle{acl_natbib}

\appendix

\clearpage
\section{Annotation Details}
\label{app:annotation}

\paragraph{Annotator backgrounds.} The 484 participants involved in our study came from 26 different countries, across Europe, Africa, Oceania, North and South America. The participants were recruited through Prolific, a crowdsourcing platform\footnote{\url{https://www.prolific.co}}. To qualify as experts, participants were required to have attained a formal education in the field and have worked in the area for at least 3 years. Participants were told that their annotations will be used to evaluate the capabilities of large language models to provide truthful answers with well-supported evidences to questions from different fields. They were also informed that the data will be released publicly upon the completion of the study.

\paragraph{Annotator fields.} The initial set of fields were listed by going through university department names, and ensuring that we cover a wide range of disciplines. Upon completing stage 1 of our annotation, we further refined these fields to represent a diverse set, for which we have enough experts.

\paragraph{Annotation costs.}  In both stage 1 and stage 2, annotators were compensated at the rate of \$15 per hour with additional bonuses when annotators spent more time than we anticipated. The average time taken for stage 2 annotations was 13.83 minutes per question-answer pair. Since this task is intensive, a single annotation task was broken down into 1-3 question-answer pairs. 

\begin{figure*}[t!]
        \begin{adjustwidth}{-1.5cm}{-1.5cm}
        \vspace{-1.7cm}
        \centering
        \scalebox{1.2}{\includegraphics[width=2\columnwidth]{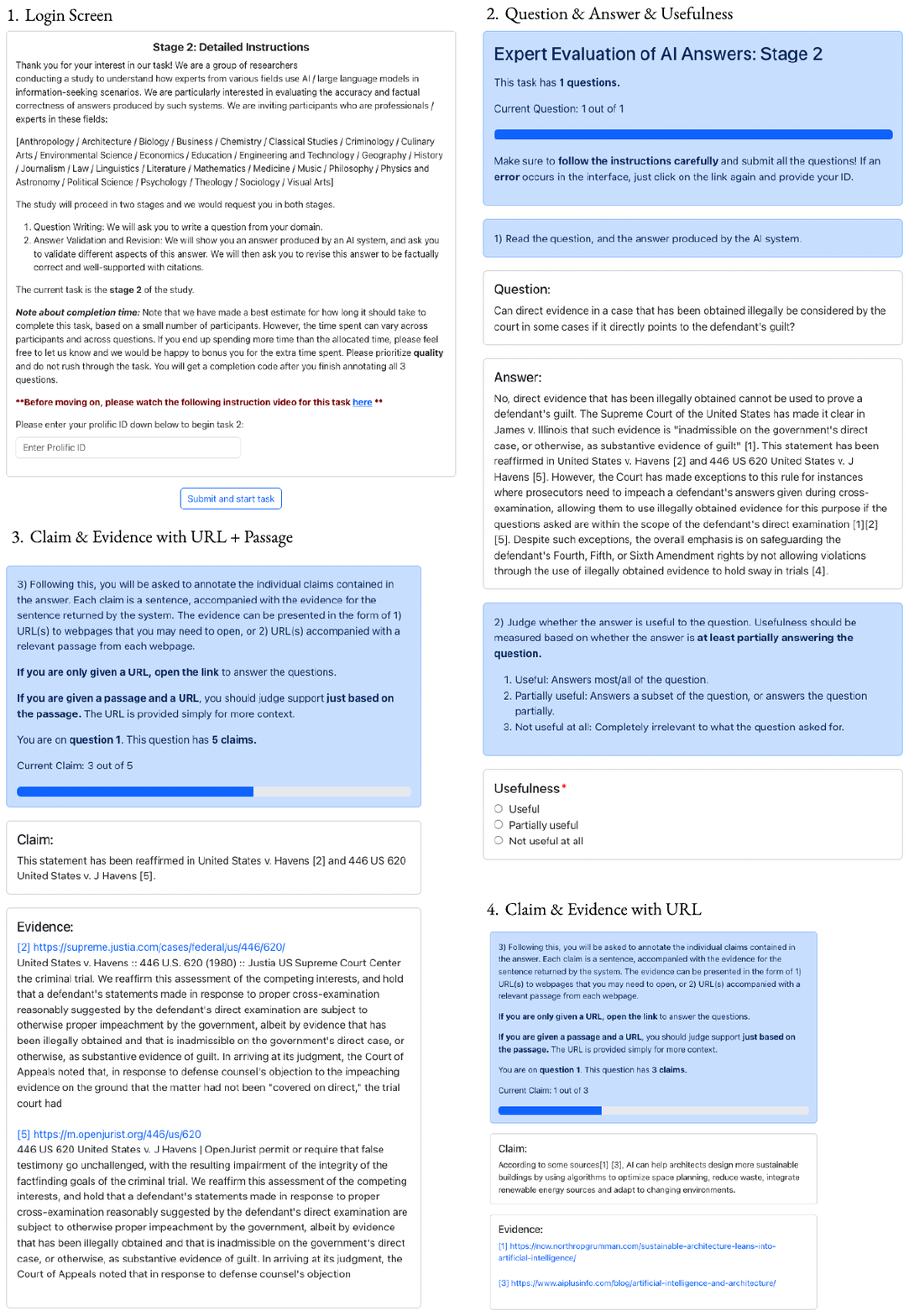}}
        \caption{Screenshots of the interface (1-4).}
        \label{fig:interface1}
    \end{adjustwidth}
\end{figure*}

\begin{figure*}[t!]
    \begin{adjustwidth}{-1.5cm}{-1.5cm}
        \vspace{-1.7cm}
        \centering
        \scalebox{1.2}{\includegraphics[width=2\columnwidth]{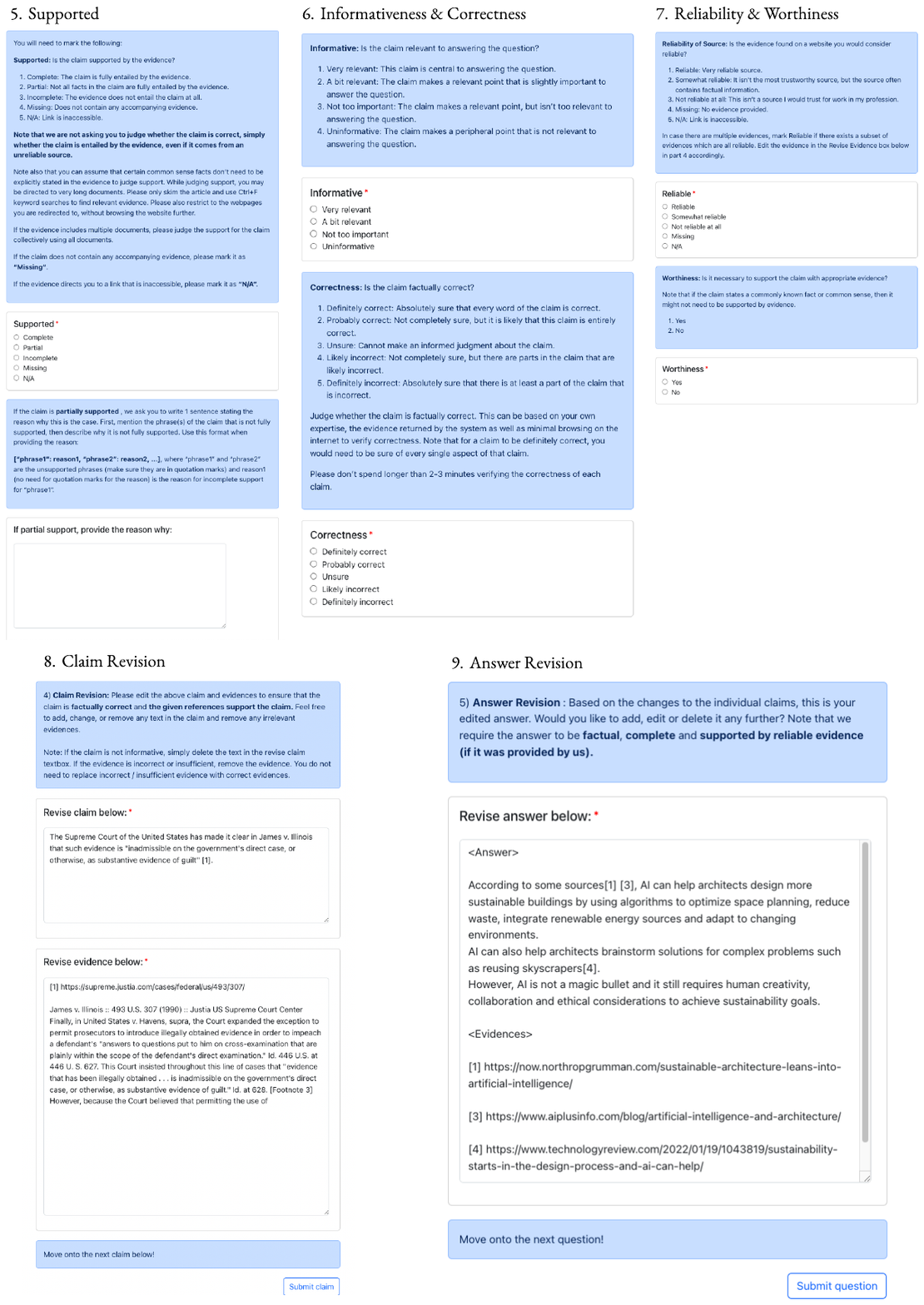}}
        \caption{Screenshots of the interface (5-9).}
        \label{fig:interface2}
    \end{adjustwidth}
\end{figure*}

\paragraph{Annotation interface.} Figures~\ref{fig:interface1} and \ref{fig:interface2} show screenshots of our stage 2 annotation interface.


\section{Experimental Details}
\label{app:experimental}

\subsection{Hyperparameter Settings}
\paragraph{Response collection.} Across all systems, for generating responses from \texttt{gpt4}, we use a temperature of 1.0, and a maximum length of 2048 tokens. For all retrieval components, we use \texttt{text-embedding-ada-002} as the embedding model. The retrieve-and-read systems first retrieve top-k (k=5) evidence passages from Sphere or top-10 Google search results using the question as the retrieval query. Google search results are split into passages of 1000 tokens with 200 tokens of overlap between subsequent chunks.

On the other hand, the post-hoc citation systems simply use the claims from \texttt{gpt4} responses, but generate their own attributions by retrieving evidence for each claim in the answer. Post-hoc retrieval systems use the top-k passages (k=5) retrieved from Sphere or the top-10 Google search results with the claim as the retrieval query. Search result are split into passages the same way as retrieve-and-read systems.

\paragraph{Automatic attribution and factuality estimation.}
For automatic attribution with AutoAIS, we use the \texttt{t5\_xxl\_true\_nli\_mixture}\footnote{\url{https://huggingface.co/google/t5_xxl_true_nli_mixture}} with 11B parameters by \citet{honovich-etal-2022-true-evaluating}. 
For finetuning the \texttt{t5\_xxl\_true\_nli\_mixture} model on the train split of \expertqa, we use the DeepSpeed ZeRO optimization \cite{rajbhandari2020zero} with stage 3, a batch size of 1, a learning rate of $1e^{-4}$ and train models for 3 epochs. 

\paragraph{Long-form QA.} For finetuning FlanT5-11B, we use a batch size of 2, maximum sequence length of 512, a learning rate of 1e-4 and train models for 3 epochs. For finetuning both Llama2-7B and Vicuna-7B, we use a batch size of 4, maximum sequence length of 2048, learning rate of $2e^{-4}$ and train models for 3 epochs.

\subsection{Prompts} The prompts used to generate responses from \texttt{gpt4} and \texttt{bing\_chat} is provided in Table~\ref{tab:qa_prompt}, while the prompt used to generate responses for retrieve-and-read systems is in Table~\ref{tab:rr_prompt}.

For factuality estimation, we use the same prompts as \citet{min2023fact} for both claim decomposition and atomic claim factuality prediction. Finally, for long-form QA baselines, we use the prompt in Table~\ref{tab:llama2_prompt} for Llama and Table~\ref{tab:vicuna_prompt} for Vicuna.
\begin{table*}[ht!]
    \centering
    \begin{tabular}{p{2\columnwidth}}
        \toprule
        \textbf{Vanilla LM QA Prompt} \\
        \midrule
        \raggedright \texttt{Answer the question completely and precisely in up to 500 words. You must provide in-line citations to each statement in the answer. The citations should appear as numbers such as [1], [2] and contain references to valid URLs on the web. A statement may need to be supported by multiple references and should then be cited as [1] [2]. \linebreak \linebreak Question: I work in the field of [FIELD]. My question is: [QUESTION] \linebreak \linebreak Answer:} \linebreak
        \hrule
    \end{tabular}
    \caption{QA Prompt for GPT4 and BingChat.}
    \label{tab:qa_prompt}
\end{table*}

\begin{table*}[ht!]
    \centering
    \begin{tabular}{p{2\columnwidth}}
        \toprule
        \textbf{Retrieve-and-read Prompt} \\
        \midrule
        \raggedright \texttt{Use the following pieces of context to answer the question completely and precisely in up to 500 words. If you don't know the answer, just say "I don't know" and explain why the context is insufficient to answer the question. \linebreak \linebreak You need to support every statement in the answer with in-line citations to passages given in the the context. The citations should appear as numbers such as [1], [2] that refer to the Passage IDs of the given passages. A statement may need to be supported by multiple references and should then be cited as [1] [2]. (for example, "Paris is the capital of France [1] [2]." where "1" and "2" are the Passage IDs of the first and second passage). \linebreak \linebreak}
        \raggedright \texttt{[CONTEXT] \linebreak \linebreak Question: [QUESTION] \linebreak Answer:}
        \linebreak
        \hrule
    \end{tabular}
    \caption{Retrieve-and-read QA prompt.}
    \label{tab:rr_prompt}
\end{table*}

\begin{table*}[ht!]
    \centering
    \begin{tabular}{p{2\columnwidth}}
        \toprule
        \textbf{Llama2 Prompt} \\
        \midrule
        \raggedright \texttt{<s>[INST] <<SYS>> \linebreak You are a helpful, respectful and honest assistant. Always answer as helpfully as possible, while being safe.  Your answers should not include any harmful, unethical, racist, sexist, toxic, dangerous, or illegal content. Please ensure that your responses are socially unbiased and positive in nature. \linebreak \linebreak If a question does not make any sense, or is not factually coherent, explain why instead of answering something not correct. If you don't know the answer to a question, please don't share false information. \linebreak <</SYS>>} \linebreak \linebreak
        \texttt{[QUESTION] [/INST]} \linebreak
        \hrule
    \end{tabular}
    \caption{Llama2 prompt for long-form QA.}
    \label{tab:llama2_prompt}
\end{table*}

\begin{table*}[ht!]
    \centering
    \begin{tabular}{p{2\columnwidth}}
        \toprule
        \textbf{Vicuna Prompt} \\
        \midrule
        \texttt{A chat between a curious user and an artificial intelligence assistant. The assistant gives helpful, detailed, and polite answers to the user's questions.} \\ \\
        \raggedright \texttt{USER: [QUESTION]\linebreak ASSISTANT:} \linebreak
        \hrule
    \end{tabular}
    \caption{Vicuna prompt for long-form QA.}
    \label{tab:vicuna_prompt}
\end{table*}

\section{Additional Plots}
\label{app:additional_res}

\begin{table*}[t]
\centering
\begin{tabular}{c|c|c|c|c|c}
    \textbf{Split} & \textbf{Model} & \textbf{R1} & \textbf{R2} & \textbf{RL} & \textbf{QFE} \\ \toprule
                            & FlanT5-11B* & 0.074 & 0.023 & 0.063 & 0.000 \\
                            & FlanT5-11B & 0.335 & 0.114 & 0.215 & 2.068 \\
                            & Vicuna-7B* & 0.358 & 0.116 & 0.209 & 0.902 \\
    \multirow{4}{*}{Random} & Vicuna-7B & 0.351 & 0.119 & 0.212 & 1.068 \\
                            & Llama2-7B* & 0.300 & 0.083 & 0.167 & 1.359 \\
                            & Llama2-7B & 0.362 & 0.125 & 0.219 & 1.985 \\
                            & Llama2-70B* & 0.320 & 0.101 & 0.181 & 1.050 \\
    \midrule
                            & FlanT5-11B* & 0.073 & 0.023 & 0.062 & 0.000 \\
                            & FlanT5-11B & 0.324 & 0.107 & 0.210 & 1.538 \\
                            & Vicuna-7B* & 0.352 & 0.114 & 0.203 & 2.596 \\
    \multirow{4}{*}{Domain} & Vicuna-7B & 0.359 & 0.120 & 0.213 & 1.739 \\
                            & Llama2-7B* & 0.303 & 0.087 & 0.169 & 1.799 \\
                            & Llama2-7B & 0.363 & 0.124 & 0.219 & 1.726 \\
                            & Llama2-70B* & 0.328 & 0.104 & 0.187 & 0.979 \\
    \bottomrule
\end{tabular}
\caption{Long-form QA results before (marked with *) and after finetuning models on the random and domain splits of \textsc{ExpertQA}.}
\vspace{-10pt}
\label{tab:lfqa_results_full}
\end{table*}

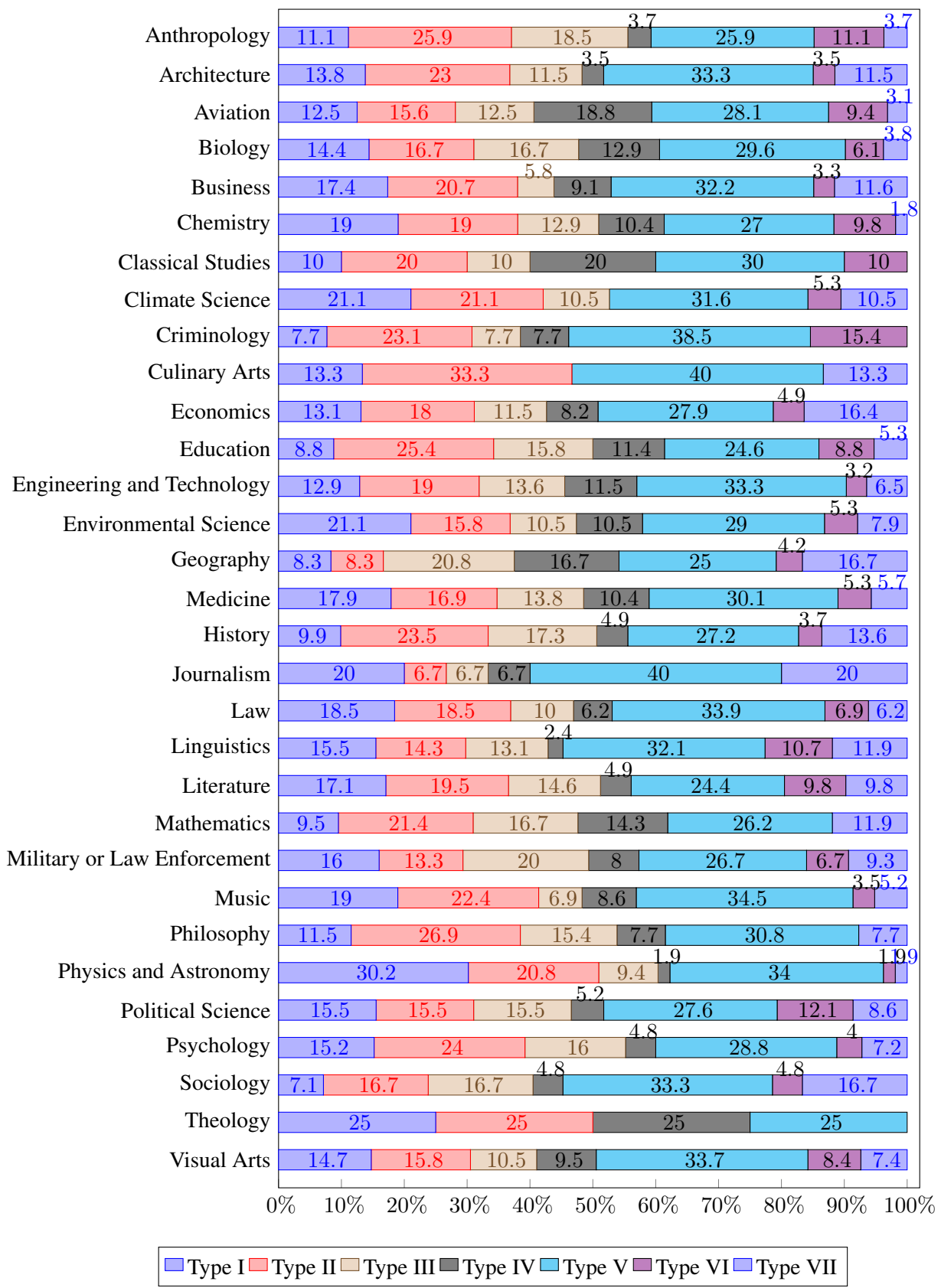
\begin{figure*}[ht]  
\centering  
\begin{tikzpicture}
  \begin{axis}[
      xbar stacked,
      width=12.6cm, height=22cm, 
      bar width=10pt,
      nodes near coords={
        \pgfkeys{/pgf/fpu=true}
        \pgfmathparse{\pgfplotspointmeta}
        $\pgfmathprintnumber[fixed, precision=1]{\pgfmathresult}$
        \pgfkeys{/pgf/fpu=false}
      },
      nodes near coords custom/.style={
        every node near coord/.style={
          check for small/.code={
            \pgfkeys{/pgf/fpu=true}
            \pgfmathparse{\pgfplotspointmeta<#1}%
            \pgfkeys{/pgf/fpu=false}
            \ifpgfmathfloatcomparison
              \pgfkeysalso{above=.05em}
            \fi
          },
          check for small,
        },
      },
      nodes near coords custom=6,
      xmin=0, xmax=102,
      xtick={0, 10, ..., 100},
      ytick={1,...,31},
      yticklabels={Visual Arts,Theology,Sociology,Psychology,Political Science,Physics and Astronomy,Philosophy,Music,Military or Law Enforcement,Mathematics,Literature,Linguistics,Law,Journalism,History,Medicine,Geography,Environmental Science,Engineering and Technology,Education,Economics,Culinary Arts,Criminology,Climate Science,Classical Studies,Chemistry,Business,Biology,Aviation,Architecture,Anthropology},
      xtick pos=bottom,
      ytick pos=left,
      xticklabel={
        \pgfkeys{/pgf/fpu=true}
        \pgfmathparse{\tick}
        $\pgfmathprintnumber[fixed, precision=1]{\pgfmathresult}\%$
        \pgfkeys{/pgf/fpu=false}
      },
      enlarge y limits=.025,
      legend style={at={(0.35,-0.05)}, anchor=north, legend columns=-1},
    ]
      \addplot coordinates {(14.74, 1) (25.0, 2) (7.14, 3) (15.2, 4) (15.52, 5) (30.19, 6) (11.54, 7) (18.97, 8) (16.0, 9) (9.52, 10) (17.07, 11) (15.48, 12) (18.46, 13) (20.0, 14) (9.88, 15) (17.89, 16) (8.33, 17) (21.05, 18) (12.9, 19) (8.77, 20) (13.11, 21) (13.33, 22) (7.69, 23) (21.05, 24) (10.0, 25) (19.02, 26) (17.36, 27) (14.39, 28) (12.5, 29) (13.79, 30) (11.11, 31)}; 
      \addplot coordinates {(15.79, 1) (25.0, 2) (16.67, 3) (24.0, 4) (15.52, 5) (20.75, 6) (26.92, 7) (22.41, 8) (13.33, 9) (21.43, 10) (19.51, 11) (14.29, 12) (18.46, 13) (6.67, 14) (23.46, 15) (16.86, 16) (8.33, 17) (15.79, 18) (19.0, 19) (25.44, 20) (18.03, 21) (33.33, 22) (23.08, 23) (21.05, 24) (20.0, 25) (19.02, 26) (20.66, 27) (16.67, 28) (15.62, 29) (22.99, 30) (25.93, 31)}; 
      \addplot coordinates {(10.53, 1) (0.0, 2) (16.67, 3) (16.0, 4) (15.52, 5) (9.43, 6) (15.38, 7) (6.9, 8) (20.0, 9) (16.67, 10) (14.63, 11) (13.1, 12) (10.0, 13) (6.67, 14) (17.28, 15) (13.78, 16) (20.83, 17) (10.53, 18) (13.62, 19) (15.79, 20) (11.48, 21) (0.0, 22) (7.69, 23) (10.53, 24) (10.0, 25) (12.88, 26) (5.79, 27) (16.67, 28) (12.5, 29) (11.49, 30) (18.52, 31)}; 
      \addplot coordinates {(9.47, 1) (25.0, 2) (4.76, 3) (4.8, 4) (5.17, 5) (1.89, 6) (7.69, 7) (8.62, 8) (8.0, 9) (14.29, 10) (4.88, 11) (2.38, 12) (6.15, 13) (6.67, 14) (4.94, 15) (10.41, 16) (16.67, 17) (10.53, 18) (11.47, 19) (11.4, 20) (8.2, 21) (0.0, 22) (7.69, 23) (0.0, 24) (20.0, 25) (10.43, 26) (9.09, 27) (12.88, 28) (18.75, 29) (3.45, 30) (3.7, 31)}; 
      \addplot[color=black, fill=cyan!50] coordinates {(33.68, 1) (25.0, 2) (33.33, 3) (28.8, 4) (27.59, 5) (33.96, 6) (30.77, 7) (34.48, 8) (26.67, 9) (26.19, 10) (24.39, 11) (32.14, 12) (33.85, 13) (40.0, 14) (27.16, 15) (30.06, 16) (25.0, 17) (28.95, 18) (33.33, 19) (24.56, 20) (27.87, 21) (40.0, 22) (38.46, 23) (31.58, 24) (30.0, 25) (26.99, 26) (32.23, 27) (29.55, 28) (28.12, 29) (33.33, 30) (25.93, 31)}; 
      \addplot[color=black, fill=violet!50] coordinates {(8.42, 1) (0.0, 2) (4.76, 3) (4.0, 4) (12.07, 5) (1.89, 6) (0.0, 7) (3.45, 8) (6.67, 9) (0.0, 10) (9.76, 11) (10.71, 12) (6.92, 13) (0.0, 14) (3.7, 15) (5.28, 16) (4.17, 17) (5.26, 18) (3.23, 19) (8.77, 20) (4.92, 21) (0.0, 22) (15.38, 23) (5.26, 24) (10.0, 25) (9.82, 26) (3.31, 27) (6.06, 28) (9.38, 29) (3.45, 30) (11.11, 31)}; 
      \addplot coordinates {(7.37, 1) (0.0, 2) (16.67, 3) (7.2, 4) (8.62, 5) (1.89, 6) (7.69, 7) (5.17, 8) (9.33, 9) (11.9, 10) (9.76, 11) (11.9, 12) (6.15, 13) (20.0, 14) (13.58, 15) (5.72, 16) (16.67, 17) (7.89, 18) (6.45, 19) (5.26, 20) (16.39, 21) (13.33, 22) (0.0, 23) (10.53, 24) (0.0, 25) (1.84, 26) (11.57, 27) (3.79, 28) (3.12, 29) (11.49, 30) (3.7, 31)}; 
      \legend{\strut Type \RNum{1}, \strut Type \RNum{2}, \strut Type \RNum{3}, \strut Type \RNum{4}, \strut Type \RNum{5}, \strut Type \RNum{6}, \strut Type \RNum{7}}
  \end{axis}
\end{tikzpicture}
\caption{The distribution of question types across all fields included in \textsc{ExpertQA}.} 
\label{fig:field_qtypes}
\end{figure*}

Examples from all fields included in \textsc{ExpertQA} are shown in Table~\ref{tab:full_examples}.
We show the distribution of all question types (from Table~\ref{tab:question_types}) across all fields that are part of \textsc{ExpertQA} in Figure~\ref{fig:field_qtypes}.

\begin{figure*}[t!]
    \centering
    \includegraphics[width=2\columnwidth]{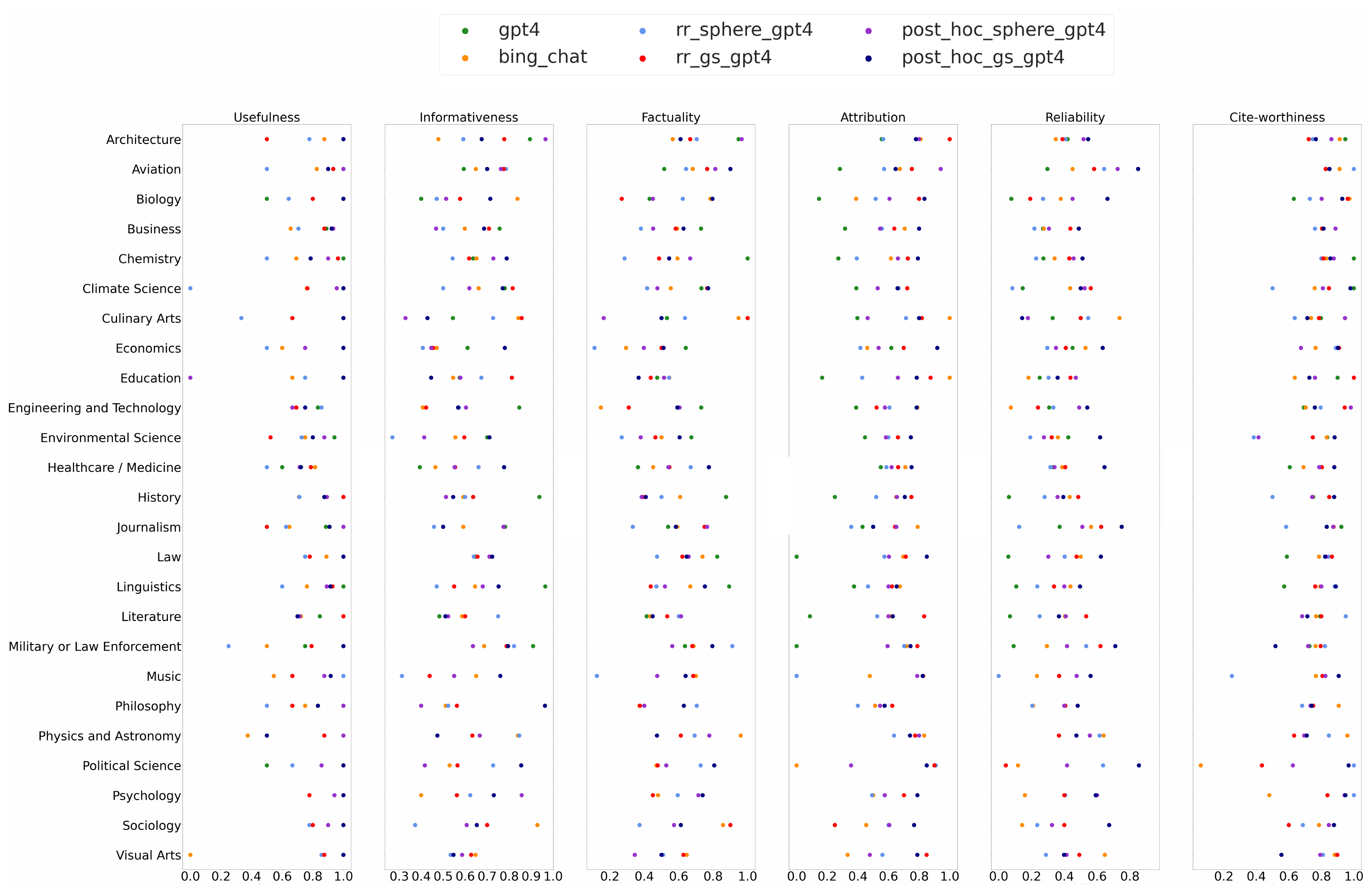}
    \caption{Label distribution of claim properties across different fields for all systems.}
    \label{fig:stats_by_field}
\end{figure*}

\begin{figure*}[t!]
    \centering
    \includegraphics[width=2\columnwidth]{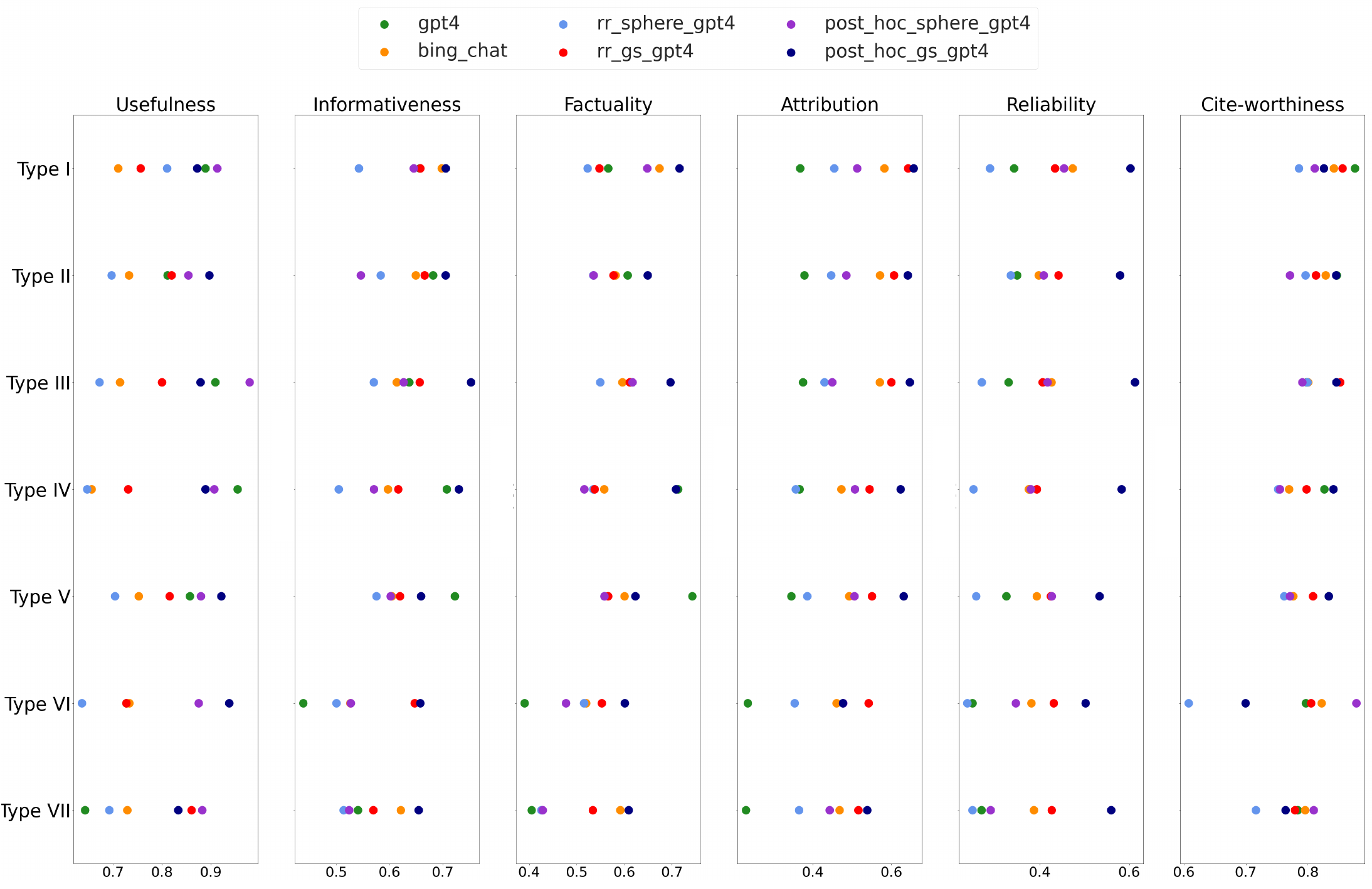}
    \caption{Label distribution of claim properties across different question types for all systems.}
    \label{fig:stats_by_qtype}
\end{figure*}

In Table~\ref{fig:stats_by_field}, we summarize the label distribution of all claim properties across fields and in Table~\ref{fig:stats_by_qtype}, we summarize the label distribution of all claim properties across question types. 

In Table~\ref{tab:lfqa_results_full}, we summarize results on long-form QA before and after finetuning models on both \textsc{ExpertQA} splits.

\begin{table*}[ht!]
\centering
\scalebox{.8}{
\rowcolors{2}{gray!15}{white}
\begin{tabular}{ c p{13cm} c }
\rowcolor{gray!40}
\textbf{Field} & \textbf{Question} & \textbf{Types} \\ \toprule
 Anthropology & \textit{Why is it that Africa's representation is still a problem in modern day times regardless of the academic writings that state otherwise?} & \RNum{2},\RNum{7} \\
 Architecture & \textit{Suppose an architect decides to reuse an existing foundation of a demolished building, what is to be considered to ensure success of the project?} & \RNum{4} \\
 Aviation & \textit{Should a low value shipment take priority from a regular customer or a high value shipment from a infrequent customer?} & \RNum{5} \\
 Biology & \textit{Can you explain the mechanisms by which habitat fragmentation affects biodiversity and ecosystem functioning, and provide examples of effective strategies for mitigating these impacts?} & \RNum{3},\RNum{6} \\
 Business & \textit{If your supplier can give you a discount for a whole yearly production, how can we take this deal without affecting our budget in a critical way?} & \RNum{5} \\
 Chemistry & \textit{Why does gallic acid have an affinity with trivalent iron ions?} & \RNum{1} \\
 Classical Studies & \textit{If researchers found a new method to unroll the Herculanum papyri, would it be fair to try it on the actual papyrus, given that it could potentially destroy it?} & \RNum{5} \\
 Climate Science & \textit{If an imidazolium based ionic liquid were to be released into the environment through the aquatic compartment, what species would be affected, if any?} & \RNum{2},\RNum{3},\RNum{5} \\
 Criminology & \textit{Mr X is an 18 year old first time offender involved in a burglary where he acted as a lookout. Which category should this information be placed under?} & \RNum{5} \\
 Culinary Arts & \textit{If mezcal production in the Valley of Mexico posits the distilling of mezcal can be traced back to ancient times, how could this be attained before the arrival of the Spaniards?} & \RNum{5} \\
 Economics & \textit{Can you summarize the current economic policies and strategies of the top five global superpowers and their potential impact on the global market?} & \RNum{1} \\
 Education & \textit{Can music therapy impact a child with autism if they have noise sensory issues?} & \RNum{5} \\
 Engineering and Technology & \textit{How different will licensing a small modular reactor be as compared to licensing traditional large nuclear power plants?} & \RNum{7} \\
 Environmental Science & \textit{Does floating solar panels minimize the risk of eutrophication or they are more trouble than their worth?} & \RNum{1} \\
 Geography & \textit{How can we overcome the limitations of remote sensing data, such as low spatial resolution and limited spectral bands?} & \RNum{4} \\
 Healthcare/Medicine & \textit{If a 48 year old woman is found to have an esophageal carcinoma that invades the muscularis propria and has regional lymph node metastases but no distant metastasis, what is her stage of cancer and what are possible recommended treatments?} & \RNum{1},\RNum{3} \\
 History & \textit{To what extent is JFK's legacy written from sympathy because of his assassination?} & \RNum{2},\RNum{7} \\
 Journalism & \textit{How many sources you must have before printing a story?} & \RNum{1} \\
 Law & \textit{Can direct evidence in a case that has been obtained illegally be considered by the court in some cases if it directly points to the defendant's guilt?} & \RNum{1} \\
 Linguistics & \textit{What are the attitudes of Received Pronunciation in the United States?} & \RNum{2} \\
 Literature & \textit{How would one go about researching the role of the mother represented in Anne Sexton's 1971 poetry volume "Transformations"?} & \RNum{4}, \RNum{6} \\
 Mathematics & \textit{Do you think there is a relation between Frobenius numbers and the Kawamata conjecture for weighted complete intersections?} & \RNum{3}, \RNum{7} \\
 Military or Law Enforcement & \textit{If you get anthrax poisoning during a mission, which chemical agent should you use to neutralise the poison?} & \RNum{1} \\
 Music & \textit{What exercises would you do in a singing class with a teenager with puberphonia?} & \RNum{4} \\
 Philosophy & \textit{How does modern neuroscience support and reject a computational theory of mind?} & \RNum{3} \\
 Physics \& Astronomy & \textit{Standard Model does not contain enough CP violating phenomena in order to explain baryon asymmetry. Suppose the existence of such phenomena. Can you propose a way to experimentally observe them?} & \RNum{5} \\
 Political Science & \textit{Despite the fact that IPCC was formed in 1988, several studies have showed that argubaly more than 50\% of all carbon emissions in history have been released since 1988. What does this show about IPCC and developed countries' efforts?} & \RNum{7} \\
 Psychology & \textit{How can counselling psychologists effectively and appropriately incorporate use of self into therapy?} & \RNum{3},\RNum{4},\RNum{7} \\
 Sociology & \textit{Which factors strengthen social cohesion within societies?} & \RNum{7} \\
 Theology & \textit{Is there any justification for the use of violence in the New Testament?} & \RNum{1} \\
 Visual Arts & \textit{Tell me the step by step process of recycling a canvas.} & \RNum{3} \\
 \bottomrule
\end{tabular}
}
\caption{Examples from \textsc{ExpertQA}, showing an example from every field included in the dataset.}
\label{tab:full_examples}
\end{table*}

\end{document}